\documentclass{article} 

\usepackage[table]{xcolor}%
\usepackage{conf,times}
\usepackage[colorlinks=true,
linkcolor=blue,
citecolor=blue,
urlcolor=blue]{hyperref}
\usepackage{url}

\usepackage{siunitx} 
\usepackage{diagbox}
\newcolumntype{x}[1]{>{\centering\arraybackslash\hspace{0pt}}p{#1}}

\usepackage{graphicx}%
\graphicspath{{Figs/}{Figs/Results/}}
\usepackage{multirow}%
\usepackage{amsmath,amssymb,amsfonts}%
\usepackage{amsthm}%
\usepackage{mathrsfs}%
\usepackage[title]{appendix}%

\usepackage{textcomp}%
\usepackage{manyfoot}%
\usepackage{booktabs}%
\usepackage{algorithm}%
\usepackage{algorithmicx}%
\usepackage{algpseudocode}%
\usepackage{listings}%

\DeclareMathOperator{\csch}{csch}
\DeclareMathOperator{\sgn}{sgn}

\usepackage{adjustbox}%
\usepackage{array}%

\newcolumntype{R}[2]{%
	>{\adjustbox{angle=#1,lap=\width-(#2)}\bgroup}%
	l%
	<{\egroup}%
}
\newcommand*\rot{\multicolumn{1}{R{15}{1em}}}

\usepackage{booktabs} 

\title{Designing a skilled soccer team for RoboCup:\\ exploring Skill-Set-Primitives \\ through reinforcement learning}

\author{Miguel Abreu \& Luis Paulo Reis  \\
	LIACC/LASI/FEUP, Artificial Intelligence \\
	and Computer Science Laboratory, \\
	Faculty of Engineering, University of Porto, \\
	Porto, Portugal \\
	\texttt{\{m.abreu,lpreis\}@fe.up.pt} \\
	\And
	Nuno Lau \\
	IEETA/LASI/DETI, Institute of Electronics \\
	and Informatics Engineering of Aveiro, \\
	Department of Electronics Telecommunications \\
	and Informatics, University of Aveiro, \\
	Aveiro, Portugal \\
	\texttt{nunolau@ua.pt}
}

\begin{document}

\maketitle
	
\begin{abstract}
The RoboCup 3D Soccer Simulation League serves as a competitive platform for showcasing innovation in autonomous humanoid robot agents through simulated soccer matches. Our team, FC Portugal, developed a new codebase from scratch in Python after RoboCup 2021. The team's performance relies on a set of skills centered around novel unifying primitives and a custom, symmetry-extended version of the Proximal Policy Optimization algorithm. Our methods have been thoroughly tested in official RoboCup matches, where FC Portugal has won the last two main competitions, in 2022 and 2023. This paper presents our training framework, as well as a timeline of skills developed using our skill-set-primitives, which considerably improve the sample efficiency and stability of skills, and motivate seamless transitions. We start with a significantly fast sprint-kick developed in 2021 and progress to the most recent skill set, including a multi-purpose omnidirectional walk, a dribble with unprecedented ball control, a solid kick, and a push skill. The push addresses low-level collision scenarios and high-level strategies to increase ball possession. We address the resource-intensive nature of this task through an innovative multi-agent learning approach. Finally, we release the team's codebase to the RoboCup community, providing other teams with a robust and modern foundation upon which they can build new features.
\end{abstract}

\section{Introduction} \label{sec:intro}

RoboCup is an international competition standing at the forefront of robotics research and innovation. It features four main leagues: \texttt{@Home} for domestic robot assistance, \texttt{Industrial} for logistics and warehousing systems, \texttt{Rescue} for disaster strategies and rescue operations, and \texttt{Soccer}, the largest league, focusing on competitive robotics.

The primary objective of this project is to enable autonomous humanoid teams to win against human soccer champions by 2050 \cite{kitano1998robocup}. Achieving this goal requires addressing the disparities between human and robotic capabilities in decision-making, coordination, and skill execution.

Our focus lies on the 3D Soccer Simulation League (3DSSL) --- a simulated soccer competition involving two teams of autonomous humanoid robots. Established in 2004, the league adopted the NAO robot model in 2008. Initially, matches were played on a 12~m by 8~m field with teams of three robots \cite{boedecker2008simspark}. In 2012, the league expanded to its current dimensions of 30~m by 20~~m, allowing for full teams of 11 players.

Some of the  3DSSL's main challenges and research efforts have been directed toward skill optimization, real-time decision-making, and cooperative multi-agent strategies \cite{gao2024survey}. Robots must execute fundamental tasks like locomotion and kicking while dynamically adapting to game scenarios involving role allocation and team coordination. Additionally, addressing these issues also requires overcoming challenges such as sensor noise, localization under uncertainty, and collision-free path planning. Some of these concerns are also applicable to real-world robotics, giving the research in this area broader implications beyond robotic soccer or RoboCup.

This paper introduces a novel approach to develop a cohesive structure of skills that works in harmony at different levels of abstraction --- from motor control to strategy --- using primitives that represent the commonalities of a skill set. All the methods detailed in this paper have been validated through participation in several official competitions, serving as the central pillar of our team, FC Portugal.

\begin{figure}[!t]
	\centering
	\includegraphics[trim={0 0 0 0}, width=0.75\columnwidth]{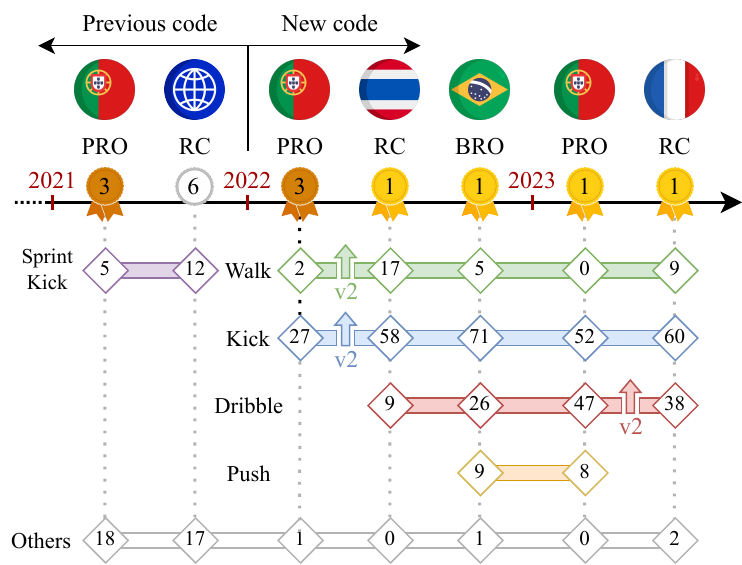}
	\caption{Timeline of skills (based on skill-set-primitives) that were used in official competitions. Each rhombus indicates the tally of goals directly attributed to each skill in the respective competition}
	\label{fig:timeline}
\end{figure}

Figure \ref{fig:timeline} depicts a timeline of skills developed with skill-set-primitives. In 2020, we developed the Sprint-Kick \footnote{First Sprint-Kick demonstration in 2020 \url{https://youtu.be/Yy1yCM5hwZI}}, a behavior that sprinted toward the ball during kickoff, resulting in 22\% of our team's goals in the Portuguese Robotics Open (PRO) 2021. In RoboCup (RC) 2021 (virtual event), the kickoff was refined, accounting for 38\% of all our goals \footnote{Successful kickoffs in RoboCup 2021 \url{https://youtu.be/3MND8RVUPBQ}}. We secured 3rd and 6th place, respectively, having good individual skills but lacking cohesion.

Post RC 2021, we rewrote the codebase from scratch in Python with three goals: reliability (minimal skill failure and collision resistance), performance (ball possession, precision, and speed), and cohesion (harmony, fast integration, and easy transitions). A new agent debuted in PRO 2022 with only 3 skills: Get Up, Omnidirectional Walk, and Kick. The walk was developed on top of a new primitive, and the kick was fixed at around 9~m, scoring 27 goals. The walk scored twice, and the remaining goal was an own goal in our favor. The team ranked 3rd, but would already win consistently against our old codebase.

For RC 2022 (Thailand), the walk and kick were upgraded, with the latter now offering distance control and a fixed long option. A secret weapon was unveil~ed later in the competition --- a pioneer and very efficient close-control dribble. The new skills along with a hand-tuned strategy led FC Portugal to its second 3DSSL victory since 2006.

Creating strategies is simpler when players can move freely to strategic positions, but it gets complex when collisions are inevitable. The Push skill was introduced to handle close encounters by using reinforcement learning for both motor control and strategy in a multi-agent environment. This configuration secured two first places in the Brazilian Robotics Open (BRO) in 2022 and PRO 2023.

The dominant dribbling skill allowed any player, even the goalkeeper, to quickly bypass all opponents and score, while keeping the ball between both feet, making defense nearly impossible without fouling. To improve gameplay, RoboCup 2023 (Bordeaux) introduced new rules to prevent ball-holding \footnote{Official rules are found at \url{https://ssim.robocup.org/3d-simulation/3d-rules/}}, rendering our Dribble and Push skills unusable. Due to time constraints, only the Dribble was retrained to comply with the new rules. FC Portugal won the main competition for the third time in 2023. 

Overall, development has primarily focused on offensive skills, with the goalkeeper serving as a strategic position player without specific skills.

While the landscape of robotic soccer teams leans toward Java or C++ programming languages, our approach leveraged the rapid development capabilities of Python, coupled with C++ modules. In this work, we also release the codebase of our team to the community, providing a foundation upon which other teams can build and innovate. The codebase includes an integrated reinforcement learning gym to facilitate the development of new skills.

In addition to releasing the codebase, this work introduces a novel learning methodology with significant contributions. We employ skill-set-primitives to capture commonalities of a set of actions, promoting seamless transitions and reducing the complexity of the learned model. These primitives allow the policy to be simplified into a shallow neural network with a single hidden layer, which improves sample efficiency and stability, while still retaining excellent performance in learning complex behaviors. In the multi-agent setting, we address the resource-intensive task of learning high-level strategies by using virtual agents and rendering only the physical robots of the closest player to the ball on each team. Furthermore, this paper showcases a collection of high-quality skills developed using the proposed learning methodology, with their performance demonstrated in official competitions. Finally, we provide comprehensive information for reproducing these skills.

\section{Background}

Typical reinforcement learning problems can be described as a Markov Decision Process (MDP) -- a tuple $\left\langle \mathcal{S,A},\Psi,p,r\right\rangle$, with a set of states $\mathcal{S}$, a set of actions $\mathcal{A}$, a set of possible state-action pairs $\Psi\subseteq \mathcal{S}\times \mathcal{A}$, a transition function $p(s,a,s'):\Psi \times \mathcal{S} \rightarrow[0,1]$, and a reward function $r(s,a):\Psi \rightarrow {\rm I\!R}$.

\subsection{MDP symmetries} \label{sec:mdp_sym}

Model reduction allows the exploitation of redundant or symmetric features. To this end, Ravindran and Barto~\cite{ravindran2001} proposed a mathematical formalism to describe MDP homomorphisms --- a transformation that groups equivalent states and actions. An MDP homomorphism $h$ from $M=\left\langle \mathcal{S,A},\Psi,p,r\right\rangle$ to $\check{M}=\left\langle \mathcal{\check{S},\check{A}},\check{\Psi},\check{p},\check{r}\right\rangle$ can be defined as a surjection $h:\Psi \rightarrow \check{\Psi}$, which is itself defined by a tuple of surjections $\left\langle f,\{g_s|s\in \mathcal{S}\} \right\rangle$. Due to symmetry characteristics, for $(s,a)\in \Psi$, $h((s,a))=(f(s),g_s(a))$ is bijective, where $f:\mathcal{S}\rightarrow \mathcal{\check{S}}$ and $g_s:\mathcal{A}_s\rightarrow \mathcal{\check{A}}_{f(s)}$, for $s\in \mathcal{S}$, satisfy:

\begin{align}
	p(f(s),g_s(a),f(s')) & = p(s,a,s'), \quad \forall s, s' \in \mathcal{S},a\in \mathcal{A}_s, \label{eq:homTran2}
	\\[3pt]
	\mathrm{and} \quad  r(f(s),g_s(a)) & =r(s,a), \quad \forall s \in \mathcal{S}, a \in \mathcal{A}_s. \label{eq:homRew2}
\end{align}

A deduction of these equations can be found in previous work \cite{abreu2023ASL}.

\subsection{Learning algorithm} \label{sec:ppopsl}

The reinforcement learning algorithm used in this work is the Proximal Policy Optimization (PPO) \cite{schulman2017ppo} extended with Proximal Symmetry Loss (PSL)\footnote{PPO+PSL algorithm: \url{https://github.com/m-abr/Adaptive-Symmetry-Learning}} \cite{abreu2023ASL}. PPO was chosen due to its good performance in single-agent and cooperative multi-agent games \cite{yu2022surprising}. PSL leverages the symmetry properties of the NAO robot (sagittal plane), to improve the training sample efficiency and also promote a more human-like behavior, without hindering asymmetric exploration. The extended objective function can be written as:

\begin{align}
	L^{PPO+PSL}(\theta, \omega)&=\hat{\mathbb{E}}_t \left[\; L_{t}^{C}(\theta) - L_{t}^{VF}(\omega) + cH(\theta) - L_{t}^{PSL}(\theta, \omega) \;\right], \\
	\text{with} \quad
	L_{t}^{C}(\theta)&=\min\left(r_t(\theta)\hat{A}_t \;,\; (1+\sgn(\hat{A}_t)\epsilon)\hat{A}_t\right), \label{eq:lclip}\\
	r_t(\theta)&=\frac{\pi_{\theta}(a_t \mid s_t)}{\pi_{\theta_{old}}(a_t \mid s_t)},\label{eq:ppo_ratio}
\end{align}

\noindent
where the stochastic policy $\pi_\theta$ is parameterized by $\theta$ and the value function by $\omega$. $\pi_{\theta_{old}}$ is a copy of the policy before each update, $\hat{A}_t$ is the estimator of the advantage function, $L_{t}^{VF}$ is a squared error loss to update the value function, $\epsilon$ is a clipping parameter, $c$ is a coefficient and $H$ is the policy's distribution entropy. The expectation $\hat{\mathbb{E}}_t$ indicates the empirical average over a finite batch of samples. The symmetry loss $L^{PSL}$ is characterized as follows:

\begin{align}
	L^{PSL}(\theta,\omega) &= \hat{\mathbb{E}}_{t} \left[\,\text{\textit{w}}_\pi \cdot L_t^{\pi}(\theta) + \text{\textit{w}}_V \cdot L_t^{V}(\omega)\,\right], \label{eq:psl}
	\\[10pt]
	\text{with} \quad
	L^{\pi}(\theta) &= -\hat{\mathbb{E}}_{t} \left[\, \min(x_t(\theta),1+\epsilon)\,\right], \label{eq:psl_pi_loss}
	\\[3pt]
	L^{V}(\omega) &= \hat{\mathbb{E}}_{t} \left[\,(V_\omega(f(s_t)) - V_t^\text{targ})^2 \,\right], 
\end{align}

\noindent
where $\textit{w}_\pi$ and $\textit{w}_V$ are weight vectors, $V_\omega(f(s_t))$ is the symmetric state value, $V_t^\text{targ}$ is the same target value used by PPO in $L^{VF}$, and $x_t(\theta)$ is a symmetry probability ratio. An extensive description of this algorithm can be found in previous work \cite{abreu2023ASL}.

\subsection{3D Simulation League} \label{sec:league}

The RoboCup 3D simulation league uses SimSpark~\cite{xu2013simspark} as its physical multi-agent simulator. The league's environment is a 30~m by 20~m soccer field containing several landmarks that can be used for self-localization: goalposts, corner flags, and lines. Each team consists of 11 humanoid robots modeled after the NAO robot, created by SoftBank Robotics. Agents get internal data (joints, accelerometer, gyroscope, and foot pressure sensors) with a one-step delay every 0.02~s, and visual data (restricted to a 120$^{\circ}$ vision cone) every 0.04~s (the visual data interval was 0.06~s prior to 2023). Agents can send joint commands every 0.02~s and message teammates every 0.04~s. In addition to the standard NAO model (type 0), there are alternative types with longer limbs (type 1), quicker feet (type 2), wider hips and longest limbs (type 3), and added toes (type 4).

For RoboCup 2023, the Technical Committee has approved a ball-holding rule to prevent the first version of FC Portugal's dribble, as it was difficult to stop without committing a charging foul. The new rule triggers a foul when a player keeps the ball within 0.12~m of itself for over 5.0~s, with no opponents closer than 0.75~m.

\section{Related Work}

RoboCup is an  international robotics competition that drives AI and robotics research, featuring standardized challenges for robotic teams. Most of the original challenges in 1997 \cite{kitano1997robocup} still stand today in the 3D soccer simulation league (3DSSL), including multi-agent collaboration, real-time reasoning, handling dynamic environments, and learning complex tasks. 

\subsection{Learning paradigms} \label{sec:learning_paradigms}

Developing a team involves the creation of an interplay of skills at different levels of abstraction, requiring seamless transitions and coordination between objectives of varying granularity. Hierarchical reinforcement learning is one way of tackling this issue by allowing the decomposition of complex tasks. In light of this, Stone et al. \cite{stone2000layered} proposed the Layered Learning (LL) paradigm, where tasks are learned in a bottom-up approach, originally in a sequential way. A lower-level layer (A) is learned and frozen before learning the next layer (B). Posterior formulations include concurrent \cite{whiteson2003concurrent} and overlapping variations \cite{macalpine2018overlapping,macalpine2015ut}. In concurrent LL, (A) is initially learned in isolation, being later refined while simultaneously learning (B). In overlapping LL, there are 3 paradigms that can be succinctly described as: the combination of independently learned layers, a partial version of concurrent LL where (A) is only partially refined, and an iterative refinement of each layer.

Still in the context of learning paradigms, it is important to address the case in which an initial model is built analytically and then refined through optimization algorithms. The most common solutions are optimizing parameters of the analytical formulation, or building a new adjustment layer that modifies the initial model's output to achieve the target behavior. The difference between the original model and the desired outcome is often called residuals. This decomposition technique is common in robotic environments, particularly when analytical models struggle to accurately represent the complex array of physical interactions. In this context, RL has been employed to refine control models, enhancing performance in manipulation and control tasks \cite{johannink2019residual, silver2018residual}. Furthermore, applying residual adjustments has been shown to improve robotic locomotion, enabling more agile and stable movements in unpredictable terrains \cite{kasaei2021cpg, kumar2022adapting, youm2023imitating}. These techniques have been widely used, ranging from dexterous hand-object interactions \cite{garcia2020physics} to improving grasping and throwing accuracy \cite{zeng2020tossingbot}, highlighting the effectiveness of residual learning in refining robotic behaviors across diverse tasks.

In a cooperative-competitive soccer environment, addressing the game's multi-agent nature is crucial, even when developing low-level skills. This is particularly important when multiple agents contend for the ball and collisions are likely to occur. Multi-agent reinforcement learning (MARL) presents three major information structures \cite{zhang2021MARL}: a centralized setting where a central controller shares information with all agents, a decentralized setting where agents can communicate with each other, and a fully decentralized setting where no explicit communication occurs. The mathematical framework underlying a 3DSSL game can be characterized as a decentralized partially observable MDP (Dec-POMDP) \cite{bernstein2002complexity}. However, during training, each agent can access all available information, making the widely-adopted centralized-learning-decentralized-execution scheme \cite{azzam2023swarm,gupta2017cooperative,lowe2017multi,simoes2020multi,smit2023scaling} a desirable option.

\subsection{Practical applications}

Gao et al. \cite{gao2024survey} present an extensive list of skills and other relevant research concerning the 3DSSL. In the scope of this work, the main skill-related contributions can be classified into three categories --- locomotion, dribbling, and kicking. For locomotion, most works focus on building a strong analytical foundation based on the Zero Moment Point (ZMP) and a model that abstracts the robot's dynamics, both historically \cite{hirai1998development,kajita2003biped} and in recent contributions \cite{kasaei2019fast,liang2014omnidirectional,seekircher2016adaptive,shafii2015learning}. One popular simplification approach is the linear inverted pendulum model (LIPM). 

The effectiveness of these models can be improved through various approaches, such as interpolation of keyframes \cite{muniz2016keyframe,shi2016adaptive,shi2016novel,Wang2015} or optimization methods. Optimization can be used to complement analytical models, either in parallel \cite{kasaei2021robust,kasaei2023learning,melo2022learning,tao2021gait} or sequentially \cite{fischer2020learning}. It can be used over primitives extracted from observed trajectories \cite{calderon2009generating} or soccer motion capture data \cite{liu2022motor}, or to learn from scratch, as demonstrated in activities like walking \cite{tao2022parallel} and running \cite{abreu2019lowlevel,abreu2019learning,melo2021learning,melo2019learning}.

Dribbling is an extension of locomotion, with the added effort of controlling the ball. This skill can be divided into close control dribble \cite{abreu2019lowlevel}, involving precise ball manipulation while keeping it near the player's feet; and speed dribble \cite{leottau2015study,leottau2015ball,muzio2022deep}, where the robot executes precise kicks to send the ball a bit further and then swiftly pursues it.

Finally, in the literature, kicking is predominantly categorized into two main forms: as the culmination of locomotion skills (implying that the robot is moving at a non-negligible speed before kicking) \cite{abreu20216d,rezaeipanah2021performing} or as a skill that involves a slower preparation stage before kicking.  In the latter case, contributions vary from attempting to cover the largest distance on a given axis \cite{depinet2015keyframe,hu2020apply,jouandeau2014optimization}, including sideways kicks \cite{dorer2018learning}, to customizing the target distance \cite{abdolmaleki2017learning,kawazoe2019design} and the kick direction \cite{spitznagel2021deep}. The static or moving initial conditions represent a trade-off between accuracy and reaction time.

\section{Skill-set-primitives} \label{sec:ss_primitives}

This section introduces a type of analytical structure that serves as the foundation for learning more complex behaviors. In this work, we propose \textit{skill-set-primitives}, a subclass of \textit{motion primitives} \cite{denk2003synthesis,nakanishi2004learning} designed for a set of skills instead of a single behavior. This type of primitive captures the commonalities of a set of actions. For instance, while walking, humans lift each foot alternately, regardless of the direction of motion, even during rotations. This underlying movement template can also be applied to marching, running, or dribbling a ball.

Skill-set-primitives are characterized by several key attributes that distinguish them within the realm of control schemes for robotic motion:

\begin{itemize}
	\item 1-to-many scheme --- instead of using primitives as building blocks to develop other skills in a 1-to-1 or many-to-1 scheme  \cite{calderon2009generating,liu2022motor}, we use them as an integrating foundation to develop and unite a specific set of skills;
	\item Continuity --- identifying a persistent pattern, often cyclic, facilitates transitions between elements within the skill set, since they share the same root. In practice, this means that the alternating foot lift primitive mentioned above can operate in the background continuously while a soccer agent seamlessly transitions between walking, running, and dribbling. In addition to providing the skill set with a uniform style, which makes transitions easier, they can also be synchronized through the primitive's cycle state.
	\item Reduced bias --- In contrast to analytical models that employ comprehensive solutions (e.g., walking engines) or primitives based on fully developed behaviors, skill-set-primitives can adopt a simpler form, as they are not required to mimic functional behaviors; they can even be abstract or unintelligible. Consequently, skill-set-primitives contain less information, leading to a reduced number of error-prone assumptions embedded in behaviors.
	\item Constant and reliable --- primitives can be designed analytically or learned as the first step of Sequential Layered Learning \cite{stone2000layered}. In both cases, using the many-to-1 scheme requires the resulting model to be frozen and provide reliable outputs for all potential input combinations. Primitives with feedback from the environment (close-loop control) are discouraged in large state spaces since they may contain unexplored areas. These areas may be visited while learning new skills, leading the primitive to produce erratic outputs that can impede the training process.
\end{itemize}

In addition to promoting skill integration due to shared motion patterns, skill-set-primitives also accelerate development by serving as a logical starting point. Reiterating, the introduced bias is reduced, as primitives do not need to mimic functional behaviors. However, providing the skill set with a uniform style can be exploited to produce symmetric and human-like movements, enhancing the aesthetics and realism of humanoid robot skills.

\section{Skill Sets}

\begin{figure}[!t]
	\centering
	\includegraphics[trim={0 0 0 0}, width=\columnwidth]{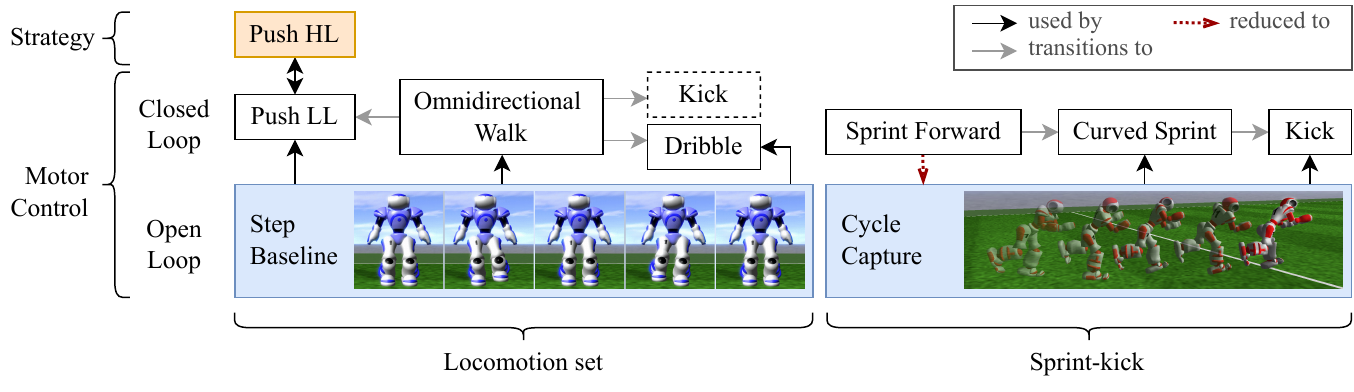}
	
	\caption{Overview of the robot's primitive-based motion framework and its hierarchical structure. The lower layer comprises skill-set-primitives (blue boxes) for two skill sets: the current locomotion set used by FC Portugal and the Sprint-Kick used in 2021. These primitives have no feedback from the environment and serve as a foundation to develop complex skills from the second layer (white boxes). The high-level Push (yellow box) is the only behavior trained in a multi-agent environment.}
	\label{fig:overview}	
\end{figure}

Two skill sets have been developed by FC Portugal: the 2021 Sprint-Kick and the more recent locomotion set. An overview of the encompassing framework can be seen in Fig.~\ref{fig:overview}. Each skill is represented by a single neural network policy, except for the locomotion set kick, which has two variants. This section will present the hierarchical structure and execution sequence of both sets, as well as the corresponding skill-set-primitives. 

\subsection{Sprint-Kick}\label{sec:sprint_kick}

The Sprint-Kick development involved three learning stages:

\begin{enumerate}
	\item \textbf{Sprint Forward}: first, the agent learned to sprint forward, with an acceleration phase followed by a cyclic phase, repeating every 14 time steps (0.28 seconds) as the robot maintained an average speed of 3.69 m/s. Average joint positions were then extracted from the cyclic phase, resulting in a skill-set-primitive with 14 hard-coded time steps;
	\item \textbf{Curved Sprint}: after accelerating, the agent learned residuals to improve the primitive, enabling it to gradually and continually change direction --- performing a curved sprint, as defined by Filter et al. \cite{filter2020new}, at an average speed of 3.44 m/s. The turning rate is a continuous parameter configurable between -33 and 33 degrees per second;
	\item \textbf{Kick}: within one meter of the ball and in the midst of sprinting, the agent learned to modify the primitive so as to collide with the ball and generate a powerful forward kick.
\end{enumerate}

\begin{figure}[!t]
	\centering
	\includegraphics[trim={0 0 0 0}, width=0.59\columnwidth]{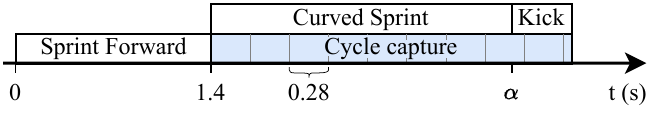}
	
	\caption{Execution of the Sprint-Kick skill set. The robot sprints forward for 1.4 seconds, then changes direction to pursue the ball, and, when closer than 1 meter ($\alpha$), it starts the kick stage.}
	\label{fig:srk_execution}	
\end{figure}

The sequence of events characterizing the Sprint-Kick is shown in Fig.~\ref{fig:srk_execution}, with $\alpha$ denoting the first moment when the sprinter is within one meter of the ball. The cycle capture primitive provides the default joint positions during the execution of the last two skills. The kick can begin from any point in the gait, as the primitive will ensure a smooth transition.

\subsection{Locomotion set}

In Section \ref{sec:ss_primitives}, we noted that the alternating lifting of each foot is a common motion trait in bipedal locomotion activities, even during ball control. To capture this trait, we simplified Kasaei et al.'s \cite{kasaei2017reliable} walking engine by fixing the model's center of mass (COM) height $c_z$ and eliminating its capacity for movement in any direction. The resulting primitive is a cyclic and smooth stationary walk, termed \textit{Step Baseline}. The equations governing the COM trajectory in $y$ ($c_y$), and the swing height ($h$) for a single step are defined as follows:

\begin{align}
	c_y(t) &= p_y\left[1 - \csch\left(\frac{T}{w}\right)\left(\sinh\left(\frac{T-t}{w}\right)+\sinh\left(\frac{t}{w}\right)\right)\right], \\
	h(t) &= H * \sin\left(\pi \cdot \frac{t}{T}\right),
\end{align}

\noindent with $w=\sqrt{c_z/g}$, where $g$ is the gravity acceleration. $T$ denotes the step duration, $t$ the elapsed time, $H$ the swing range, and $p_y$ the zero moment point's (ZMP) location in $y$ (corresponding to the position of the support foot). In the final primitive, every parameter was optimized concurrently with the Omnidirectional Walk model to maximize the walk stability and speed. Two parameter schemes were tested: static and dynamic (i.e., controlled by the walk policy). Ultimately, the static approach exhibited greater stability and allowed for easier transitions.

The skill-set-primitive was then frozen, and all parameters were reused for dribbling and pushing, except for $p_y$, which was increased to motivate a wider leg stance, except for robot R3 due to its default wide stance. Refer to \ref{app:parameters}, Table \ref{tab:step_baseline_params} for the optimized values. Note that $c_z$ is given as a percentage of the COM's height when the robot is standing with straight legs, since different robot types vary in height and weight distribution. The second dribbling version reverts to the $p_y$ used during walking, as a wider leg stance is more likely to trigger ball-holding fouls under the new rules.

\begin{figure}[!t]
	\centering
	\includegraphics[trim={0 0 0 0}, width=0.90\columnwidth]{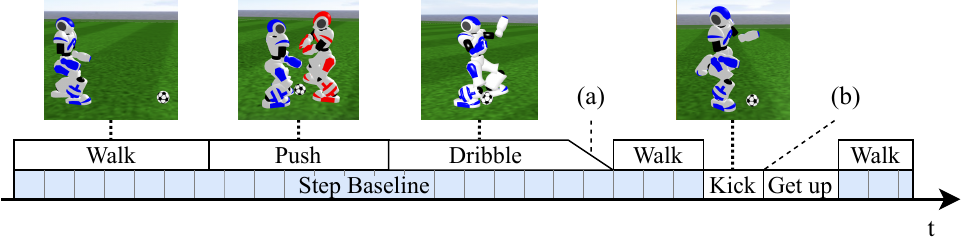}
	
	\caption{Example of a locomotion set execution sequence. The Step Baseline operates in the background as the agent walks to the ball, pushes it past the opponent, dribbles to an empty space, gradually reverts to the Step Baseline to transition to walking (\textbf{a}), and positions itself for the kick. It then performs a long kick and falls (\textbf{b}), gets up, and resumes walking.}
	\label{fig:locomotion_execution}	
\end{figure}

The locomotion set consists of the following skills, ordered chronologically:

\begin{enumerate}
	\item \textbf{Omnidirectional Walk}: the policy directs the robot to the target location while ensuring it aligns with the desired orientation. A single neural network automatically handles acceleration, deceleration, rotation, and walking in any direction;
	\item \textbf{Kick}: the Kick starts directly from the walking skill when the robot is roughly aligned with the target. The model learned to compensate for small target alignment deviations. It comprises a short kick for precise distances ranging from 3~m to 9~m and a long kick capable of reaching up to 21~m with very low lateral deviation.
	\item \textbf{Dribble}: dribbling uses the Step Baseline, and transitions directly from walking. The policy combines forward locomotion with rotation, while keeping the ball within close range. The only design distinction in the second version developed for RoboCup 2023 is its expanded ball range to comply with the new ball-holding rules. This adjustment was achieved during training by setting the reward to zero when the ball was within 0.115~meters.
	\item \textbf{Push}: the Push is comparable to dribbling but is designed for collision-prone scenarios. It comprises two concurrently learned skills. The high-level (HL) part, operating at 3.125~Hz, generates a target push direction to maximize ball possession while moving it to a user-requested location. It takes into consideration the positions of close teammates and opponents. The low-level (LL) part controls the robot joints at 50~Hz to steer the ball toward the HL target.
\end{enumerate}

Fig.~\ref{fig:locomotion_execution} presents an execution sequence example, where the Step Baseline is active during walking, pushing, and dribbling. A matrix of skill transitions is provided in \ref{app:parameters}, Table \ref{tab:transitions}. Transitions were either \textit{trained} as part of skill optimization, \textit{innate}, if they emerged naturally, or \textit{assisted}, when a gradual return to the Step Baseline is required before walking (as seen in Fig.~\ref{fig:locomotion_execution}, (\textbf{a})). 

Most skills were trained following a walking phase. The transition from Push to Dribble is innate due to their similarity, especially considering that the former skill is slower. Another instance is when the robot is still standing after kicking and subsequently starts walking.

\section{Training Environment}

This section describes the most important characteristics of the environments used to train the aforementioned skills. For a comprehensive list of environment characteristics, please refer to \ref{app:Sprint-Kick} and \ref{app:locomotion_set}, which cover details such as state space, symmetry operations (referenced with index/multiplier notation \cite{abreu2023ASL}), action space modifications, episode setup, reward functions, and hyperparameters.

\subsection{Sprint-Kick} \label{sec:train_env_sprint_kick}

\begin{figure}[!t]
	\centering
	\includegraphics[trim={0 0 0 0}, width=0.98\columnwidth]{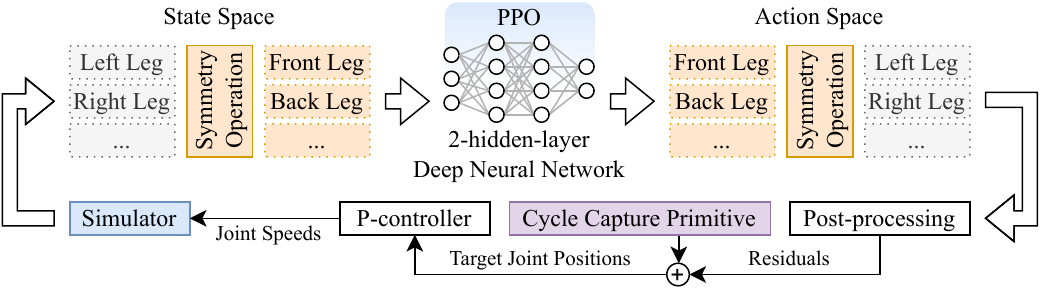}
	\caption{Sprint-Kick training environment. Symmetry operations abstract the RL algorithm from the left and right sides of the robot, forcing the policy to learn a symmetric behavior. The cycle capture primitive is extracted from \textit{Sprint Forward} and added to \textit{Curved Sprint} and \textit{Kick}.}
	\label{fig:env_srk}	
\end{figure}

The Sprint-Kick training environment is outlined in Fig.~\ref{fig:env_srk}. A 2-hidden-layer deep neural network is trained with PPO and abstracted from the left and right sides of the robot. To do this, we relabel states and actions so that the learning policy is only aware of a front and a back leg. The robot begins sprinting with the left leg in the front, and whenever the right leg takes the lead, all observations and actions are mirrored along the robot's sagittal plane. The resulting action goes through a post-processing stage including a low-pass filter, and mapping raw joint values to their respective operational ranges. The primitive that captures the \textit{Sprint Forward} motion cycle is then added to the subsequent \textit{Curved Sprint} and \textit{Kick} skills.

\subsection{Locomotion set}

\begin{figure}[!t]
	\centering
	\includegraphics[trim={0 0 0 0}, width=0.98\columnwidth]{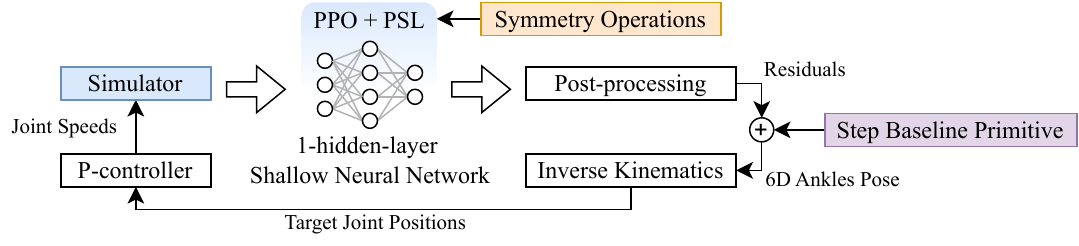}
	\caption{Training environment for the Omnidirectional Walk, Push LL, and Dribble skills, using the Step Baseline primitive. The PPO algorithm is extended with the Proximal Symmetry Loss to allow for a simpler symmetry configuration and reduce human-induced bias.}
	\label{fig:env_step}	
\end{figure}

Omnidirectional Walk, Push LL, and Dribble are skills from the locomotion set that rely on the Step Baseline primitive. Fig.~\ref{fig:env_step} shows the corresponding training environment. PPO is used in conjunction with Proximal Symmetry Loss (PSL) \cite{abreu2023ASL}, which transfers the symmetry burden to the neural network, eliminating the need for symmetry operations after training. Moreover, PSL reduces human bias by allowing the policy to explore asymmetric solutions, and it does not impose user-defined symmetry switches.

Instead of generating raw joint values for the legs, the new training environment produces relative 6D poses for both ankles. Then, a low-pass filter with a strong smoothing factor stabilizes the movement. The action values are individually mapped to the desired initial exploration range. Subsequently, the Step Baseline primitive is integrated, and the action is converted into target joint positions using inverse kinematics. These characteristics, associated with a rich state space allow the policy to be simplified into a shallow neural network with a single hidden layer with 64 neurons. This modification drastically increases sample efficiency during training, consequently reducing the learning time, while achieving excellent performance, as is demonstrated in Section \ref{sec:results}.

\subsubsection{Push HL} \label{sec:push_hl}

\begin{figure}[!t]
	\centering
	\includegraphics[trim={0 0 0 0}, width=1\columnwidth]{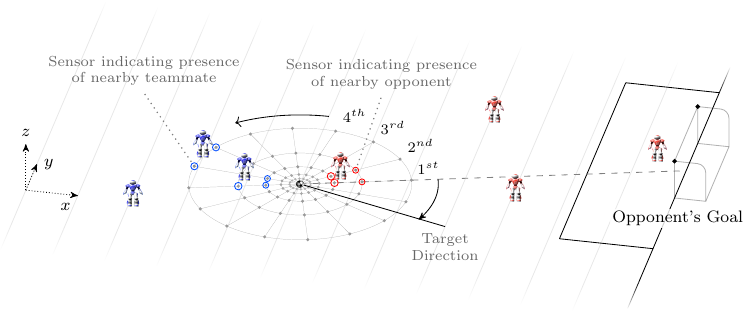}
	\caption{Push HL multi-agent training environment. The policy is aware of teammates and opponents that surround the ball when deciding a short-term target direction.}
	\label{fig:push}	
\end{figure}

The high-level part of the Push skill is trained in a multi-agent environment, as illustrated in Fig.~\ref{fig:push}. The neural network receives radar data as input, i.e., the space around the ball is segmented by 5 circles and 16 radial line segments, yielding 80 intersection points. Each point corresponds to the location of two sensors, one for detecting nearby teammates, and another for opponents, totaling 160 independent sensors. For a more interactive explanation of this training environment, please refer to the video demonstration at \url{https://youtu.be/rGWN83FBdJ4}.

The radar is always oriented toward the user-defined long-term objective direction, which corresponds to the opponent's goal during training. As shown in Fig.~\ref{fig:push}, the first sensor of the array always captures the players that are between the ball and the objective. While the order of each sensor is not important, it is imperative to the neural network that each input has a constant meaning, since it will output a short-term target direction, which is given in relation to the direction of the long-term target.

Since this training is very resource-intensive, only the closest player to the ball per team is simulated. The remaining players are represented by a 2D point with a position, velocity, and acceleration, and follow a simplified locomotion model. Despite the speed optimizations, a complete training session takes a full week to run on an Intel Xeon 6258R.

This problem fits into the centralized-learning-decentralized-execution scheme. Although the game is defined as a decentralized partially observable MDP (Dec-POMDP), the learning environment has access to the exact position of each agent and the ball. During actual games, FC Portugal uses the communication channel to address partial observability by attempting to build a common worldview. Consequently, the learning problem can be approximated and solved as a MDP. The policy primarily focuses on the ball, and the episode ends only when the ball is relocated to the center of the field. Meanwhile, the policy observes multiple teammates trying to advance the ball through their Push LL skill, which evolves over time, adding a dynamic component to the environment. The target decision is influenced by all agents that are near the ball, as well as the evolving abilities of the shared Push~LL skill.

\section{Results and Discussion} \label{sec:results}

Demonstrations of all skills can be found on YouTube \footnote{\url{https://www.youtube.com/playlist?list=PLIeNX3I5JIATFlOXfM93d6cijDdRheyyO} For a comprehensive list of available videos, refer to Appendix \ref{app:videos}.}. Major results are summarized in Table~\ref{tab:results}. The Sprint Forward skill, which is the base of the Sprint-Kick, attains an average speed of 3.69 m/s after stabilizing --- an improvement over prior work~\cite{abreu2019learning} (2.5 m/s). While Melo et al.~\cite{melo2021learning}, reported an average speed of 3.76 m/s under controlled conditions, our sprint was the fastest sprint to be integrated into a team and demonstrated in an official RoboCup competition. The Sprint-Kick sacrifices some speed for maneuverability, being ultimately used to score goals during kickoff set plays. It was also the first kick in the league to be performed while running. For more results, see Appendix \ref{app:results}.

\begin{table}[h]
	\caption{Results per skill}
	\label{tab:results}
	\begin{tabular*}{\textwidth}{@{\extracolsep\fill}lclc}
		\toprule%
		\multicolumn{2}{@{}c@{}}{Skill} & Metric &Result \\ \midrule
		\multirow{1}{*}[-6.8mm]{Sprint-Kick} & Sprint Forward & average speed & 3.69 m/s \\ \cmidrule{2-4} 
		& \multirow{2}{*}{Curved Sprint} & average speed & 3.44 m/s \\
		&  & turning radius & 6.02 m \\ \cmidrule{2-4} 
		& Kick & average distance & 9.07 m \\ \midrule
		Omnidirectional Walk & & maximum speed & 0.70\(-\)0.90 m/s \footnotemark[1] \\ \midrule
		& Short Kick & avg. error (3/6/9 m) & 0.14/0.25/0.62 m \\ \cmidrule{2-4} 
		Kick & \multirow{2}{*}{Long Kick} & average error & 1.85 m \\ 
		&  & average dist. (x) & 19.2 m \\ \midrule
		\multirow{1}{*}[-6.3mm]{Dribble} & \multirow{2}{*}{v1} & average fwd. speed & 1.25\(-\)1.41 m/s \footnotemark[2]\\ 
		&  & turn around offset & 0.8 m \\ \cmidrule{2-4} 
		& \multirow{2}{*}{v2} & average fwd. speed & 0.71 m/s \\ 
		&  & turn around offset & 0.5 m \\ \midrule
		Push & & maximum speed & 0.55 m/s \\ \bottomrule
	\end{tabular*}
	
	\footnotetext[1]{depends on robot type and walking direction}
	\footnotetext[2]{depends on robot type}
\end{table}

The Omnidirectional Walk was trained individually for each robot type, offering fast forward walking, rotating while walking in any direction, and precise approaches to kick the ball or wait in strategic positions. Maximum speeds range from 0.70 to 0.90~m/s, depending on walk direction and robot type. Comparing walking skills is challenging due to high variance in versatility. In 2015, Wang et al. \cite{Wang2015} achieved 0.5~m/s for forward and backward walking, and 0.4 m/s for side walking, while retaining omnidirectional abilities. In the same year, Shafii et al. \cite{shafii2015learning} optimized a forward walk (0.78 m/s) and a side walk (0.48 m/s). It would be unfair to compare these speeds with an omnidirectional walk. Recent advancements, such as Kasaei et al.'s \cite{kasaei2021robust} forward walk with rotation (0.956 m/s) and Fischer et al.'s \cite{fischer2020learning} specialization for robot type 4 (toes) with speeds of 1.3 m/s forward and 1.03 m/s backward, concern specific conditions. However, these approaches lack the all-encompassing control of an omnidirectional walk, with the ability to accelerate, decelerate, rotate, and change walking direction seamlessly and automatically in a single skill. Our behavior is best analyzed in action \footnote{Omnidirectional Walk training environment and demonstration: \url{https://youtu.be/dXzIuZlOFZc}}.

\begin{figure}[!t]
	\centering
	\includegraphics[trim={0 0 0 0}, width=1\columnwidth]{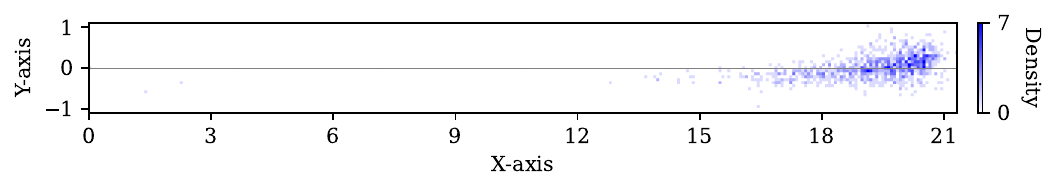}
	
	\caption{Heatmap of final ball positions after 1000 long kicks using robot type 1}
	\label{fig:long_kick_heatmap}	
\end{figure}

The Short Kick offers variable distance but was tested at 3, 6, and 9 meters, with 1000 samples, to measure distance errors (see Table~\ref{tab:results}). The Long Kick originated a heatmap with a spatial distribution of final ball positions using robot type 1 (see Fig.~\ref{fig:long_kick_heatmap}). Reaching 20 m \cite{depinet2015keyframe} or more \cite{hu2020apply} is not uncommon in the league. Moreover, some kicks have special features such as multi-directional targets \cite{spitznagel2021deep}, kick while moving  \cite{abreu20216d,rezaeipanah2021performing}, or handling fast-moving balls, which magmaOffenburg mastered for the RoboCup 2023 Kick Rolling Ball Challenge. However, success depends on factors such as kick preparation quality and repeatability, duration, accuracy, and adaptability to in-game disruptions.

In previous work \cite{abreu2019lowlevel}, we introduced a dribble developed from scratch through RL, where the ball is carried between the robot's legs at 0.31 m/s, which was never integrated into the team for being slow. Dribble v1 was the first close control dribble to be introduced in the 3D Soccer Simulation League, reaching 1.25 to 1.41 m/s, depending on the robot type. The second version complies with new rules by keeping the ball further than 12 cm, albeit at the cost of speed\footnote{Dribble evolution, training environment, and demonstration: \url{https://youtu.be/8UED_Zl-nbQ}}. Other dribble behaviors are not comparable \cite{leottau2015study,leottau2015ball,muzio2022deep}, as they focus on kicking the ball further and rapidly pursuing it, rather than maintaining close control throughout.

The Push behavior, designed for ball possession and stability in collision-prone game scenarios, reaches 0.55 m/s. In the 3D league, applying machine learning to high-level strategies remains uncommon due to the complexity of the humanoid soccer multi-agent environment. Related prior work includes Muzio et al.'s dribble technique \cite{muzio2020deep}, which trained a higher-level policy for controlling a walking engine by providing translation and rotation commands every 5th step. Despite comprising two layers, the more abstract layer does not address strategy concerns.

In contrast to Austin Villa \cite{macalpine2018overlapping}, our focus was on producing fewer behaviors to simplify maintenance. Our proposed methodology effectively handles the added complexity of developing more versatile skills. The most effective evaluation of these skills and their seamless integration is through RoboCup competitions, where FC Portugal has achieved excellent results.

\section{Locomotion Performance in RoboCup}

In the previous section, we presented results for individual skills tested in controlled conditions. While kicks are relatively straightforward to compare with published data and are often used similarly in training and competition, locomotion is heavily influenced by team strategies and must adapt to unforeseen circumstances during competitions. Moreover, comparing locomotion with published data is challenging due to the high variance in target versatility and performance evaluation metrics. To address these challenges, this section presents a statistical analysis of locomotion performance in past RoboCup competitions.

\subsection{Methodology}

As highlighted in Section~\ref{sec:intro}, the NAO robot has been the standard platform in RoboCup Soccer since 2008. Consequently, our analysis begins with data from 2008, encompassing all major RoboCup competitions through 2023. Match data were sourced from the \textit{RoboCup.info} online archive~\cite{Glaser_Archive}, and we included all rounds of each competition, including seeding rounds. Penalty shootouts, FatProxy matches, challenges, and test games were excluded. For each match, only data from the \texttt{PlayOn} segments was analyzed, as these periods impose fewer restrictions on player movement.

To ensure consistent and reliable data, we analyzed robot movement using 1-second measurement windows, which mitigates the impact of erratic movements or collisions. Specific exclusion criteria were applied to ensure the validity of these samples:

\begin{itemize}
	\item \textbf{Abnormal height:} Samples were excluded if the robot's z-axis height was below 0.25~m or above 0.40~m.
	\item \textbf{Excessive speed:} Samples were excluded if the robot's speed between consecutive time steps exceeded 4~m/s, a threshold slightly higher than the maximum running speed.
\end{itemize}

Exclusion zones were extended 1 second prior to an invalid time step to account for events like falling, which could affect speed calculations. For example, if a robot begins to fall but is beamed by the referee before its height drops below 0.25~m --- due to a play mode change or foul --- the fall itself may go undetected, yet the initial falling motion would still be incorrectly factored into speed calculations.

\subsection{Results}

Table~\ref{tab:teams_speed} summarizes the average and maximum speeds of RoboCup teams since 2008, aggregated by competition. Teams active since 2021 are ordered by the number of participations, with winners highlighted in red\footnote{For completeness, SEU-RedSun won in both 2008 and 2009. In 2008, their average and maximum speeds were 0.12~m/s and 0.61~m/s, with a 15\% fallen time and 41~m of average distance per player. In 2009, these values improved to 0.16~m/s, 0.73~m/s, 15\%, and 56~m, respectively.}. The key metrics were obtained as follows:

\begin{itemize}
	\item \textbf{Average speed:} computed by aggregating valid 1-second window samples for all teammates in a match, summing the distances covered, and dividing by the total window durations. This method is preferred over averaging individual speeds, as it gives more weight to players with more valid samples, preventing those who frequently fell or crashed from skewing the results. The process is repeated across all games and averaged per team.
	\item \textbf{Maximum speed:} determined by selecting the 5th highest speed from all valid samples across all teammates in a match, ensuring the value reflects sustained performance rather than statistical anomalies.
\end{itemize}

From preliminary analysis, team speed appears positively correlated with competition success. This observation will be explored in greater detail later in this section.

\begin{table}[t]
	\caption{Average speed (m/s) and maximum speed (m/s) aggregated by team since 2008. Teams active since 2021 are ordered by number of participations; winners highlighted in red.}
	\label{tab:teams_speed}
	\tiny
	\setlength\tabcolsep{1pt} 
	\begin{tabular*}{\textwidth}{@{\extracolsep\fill}ccccccccccccc}
		\toprule%
		& \rot{\textbf{FC Portugal}} & \rot{\textbf{magmaOffenburg}} & \rot{\textbf{UT Austin Villa}} & \rot{\textbf{BahiaRT}} & \rot{\textbf{HfutEngine3D}} & \rot{\textbf{ITAndroids}} & \rot{\textbf{Apollo3D}} & \rot{\textbf{Miracle3D}} & \rot{\textbf{KgpKubs}} & \rot{\textbf{WITS FC}} & \rot{\textbf{WrightOcean}} & \rot{\textbf{MIRG}} \\ \midrule
		\textbf{2008} & .03\, 0.28 & - & .02\, 0.17 & - & .07\, 0.44 & - & .11\, 0.81 & - & - & - & - & - \\ 
		\textbf{2009} & .09\, 0.76 & .05\, 0.26 & - & .02\, 0.10 & .15\, 0.65 & - & - & - & - & - & - & - \\ 
		\textbf{2010} & .19\, 0.84 & .14\, 0.46 & .12\, 0.75 & .03\, 0.37 & .18\, 0.60 & - & \textcolor{red}{.23\, 0.82} & - & - & - & - & - \\ 
		\textbf{2011} & .25\, 0.83 & .16\, 0.67 & \textcolor{red}{.26\, 0.74} & .05\, 0.54 & .17\, 0.73 & - & .28\, 0.86 & - & - & - & - & - \\ 
		\textbf{2012} & .25\, 0.84 & .20\, 0.73 & \textcolor{red}{.25\, 0.77} & - & - & - & .33\, 0.93 & - & - & - & - & - \\ 
		\textbf{2013} & .35\, 0.73 & .28\, 0.87 & .30\, 0.95 & .11\, 0.45 & .20\, 0.72 & .19\, 0.71 & \textcolor{red}{.37\, 0.94} & - & - & - & - & - \\ 
		\textbf{2014} & .38\, 0.78 & .35\, 0.86 & \textcolor{red}{.35\, 0.97} & .31\, 0.86 & .17\, 0.92 & - & - & - & - & - & - & - \\ 
		\textbf{2015} & .38\, 0.85 & .35\, 0.85 & \textcolor{red}{.38\, 0.97} & .36\, 0.77 & .16\, 0.92 & .26\, 0.63 & .38\, 0.95 & .33\, 0.61 & - & - & - & - \\ 
		\textbf{2016} & .39\, 0.87 & .37\, 0.85 & \textcolor{red}{.44\, 0.97} & .34\, 0.74 & .20\, 0.86 & .27\, 0.76 & - & .35\, 0.61 & .28\, 0.67 & - & - & - \\ 
		\textbf{2017} & .32\, 0.88 & .39\, 0.86 & \textcolor{red}{.45\, 0.98} & .39\, 0.87 & .36\, 0.92 & .31\, 0.70 & - & .20\, 0.84 & .23\, 0.68 & - & - & - \\ 
		\textbf{2018} & .36\, 0.88 & .38\, 1.05 & \textcolor{red}{.41\, 0.97} & .39\, 0.85 & - & .31\, 0.70 & - & .21\, 0.78 & .24\, 0.67 & - & - & - \\ 
		\textbf{2019} & .34\, 2.62 & .42\, 1.06 & \textcolor{red}{.44\, 0.97} & .41\, 0.84 & .33\, 0.83 & .34\, 0.70 & - & - & - & - & .43\, 0.89 & - \\ 
		\textbf{2021} & .38\, 3.44 & .44\, 1.07 & \textcolor{red}{.47\, 0.98} & .37\, 0.85 & .41\, 0.91 & .33\, 0.70 & .42\, 1.27 & .41\, 0.79 & .26\, 0.73 & .27\, 0.78 & .47\, 0.90 & .27\, 0.78 \\ 
		\textbf{2022} & \textcolor{red}{.48\, 1.43} & .39\, 2.46 & .47\, 0.98 & .37\, 0.95 & .40\, 0.92 & .33\, 0.71 & .43\, 0.82 & .40\, 0.79 & .32\, 0.84 & .29\, 0.78 & - & - \\ 
		\textbf{2023} & \textcolor{red}{.46\, 0.93} & .39\, 2.55 & .31\, 1.64 & .38\, 0.88 & - & .34\, 0.70 & - & - & - & .28\, 0.83 & - & - \\ \bottomrule
		
	\end{tabular*}
	
	\bigskip

	\caption{Fallen time (\%) and average distance per player (m) aggregated by team since 2008. Teams active since 2021 are ordered by number of participations; winners highlighted in red.}
	\label{tab:teams_stability}
	\tiny
	\setlength\tabcolsep{1pt} 
	\begin{tabular*}{\textwidth}{@{\extracolsep\fill}ccccccccccccc}
		\toprule%
		& \rot{\textbf{FC Portugal}} & \rot{\textbf{magmaOffenburg}} & \rot{\textbf{UT Austin Villa}} & \rot{\textbf{BahiaRT}} & \rot{\textbf{HfutEngine3D}} & \rot{\textbf{ITAndroids}} & \rot{\textbf{Apollo3D}} & \rot{\textbf{Miracle3D}} & \rot{\textbf{KgpKubs}} & \rot{\textbf{WITS FC}} & \rot{\textbf{WrightOcean}} & \rot{\textbf{MIRG}} \\ \midrule
		\textbf{2008} & 10\%\! 13\phantom{0} & - & 22\%\! 8\phantom{0}\phantom{0} & - & 27\%\! 25\phantom{0} & - & 14\%\! 41\phantom{0} & - & - & - & - & - \\ 
		\textbf{2009} & 32\%\! 27\phantom{0} & \phantom{0}6\%\! 21\phantom{0} & - & 27\%\! 7\phantom{0}\phantom{0} & 13\%\! 52\phantom{0} & - & - & - & - & - & - & - \\ 
		\textbf{2010} & \phantom{0}9\%\! 88\phantom{0} & 12\%\! 63\phantom{0} & \phantom{0}6\%\! 65\phantom{0} & 33\%\! 11\phantom{0} & \phantom{0}8\%\! 86\phantom{0} & - & \textcolor{red}{11\%\! 111} & - & - & - & - & - \\ 
		\textbf{2011} & 11\%\! 113 & \phantom{0}5\%\! 80\phantom{0} & \phantom{0}\textcolor{red}{5\%\! 122} & 27\%\! 15\phantom{0} & \phantom{0}8\%\! 76\phantom{0} & - & \phantom{0}8\%\! 128 & - & - & - & - & - \\ 
		\textbf{2012} & 10\%\! 111 & \phantom{0}3\%\! 101 & \textcolor{red}{10\%\! 114} & - & - & - & 21\%\! 124 & - & - & - & - & - \\ 
		\textbf{2013} & \phantom{0}8\%\! 159 & \phantom{0}6\%\! 136 & \phantom{0}4\%\! 143 & \phantom{0}9\%\! 48\phantom{0} & \phantom{0}5\%\! 96\phantom{0} & 14\%\! 75\phantom{0} & \phantom{0}\textcolor{red}{7\%\! 171} & - & - & - & - & - \\ 
		\textbf{2014} & \phantom{0}8\%\! 193 & \phantom{0}5\%\! 155 & \phantom{0}\textcolor{red}{6\%\! 144} & \phantom{0}7\%\! 169 & \phantom{0}8\%\! 67\phantom{0} & - & - & - & - & - & - & - \\ 
		\textbf{2015} & \phantom{0}6\%\! 169 & \phantom{0}4\%\! 168 & \phantom{0}\textcolor{red}{3\%\! 172} & \phantom{0}4\%\! 163 & \phantom{0}9\%\! 65\phantom{0} & 13\%\! 100 & \phantom{0}9\%\! 185 & 19\%\! 118 & - & - & - & - \\ 
		\textbf{2016} & \phantom{0}5\%\! 174 & \phantom{0}3\%\! 180 & \phantom{0}\textcolor{red}{3\%\! 193} & \phantom{0}4\%\! 168 & \phantom{0}9\%\! 83\phantom{0} & \phantom{0}6\%\! 123 & - & 10\%\! 155 & \phantom{0}4\%\! 139 & - & - & - \\ 
		\textbf{2017} & \phantom{0}5\%\! 156 & \phantom{0}3\%\! 199 & \phantom{0}\textcolor{red}{3\%\! 194} & \phantom{0}6\%\! 185 & 14\%\! 129 & \phantom{0}6\%\! 154 & - & \phantom{0}4\%\! 97\phantom{0} & \phantom{0}4\%\! 119 & - & - & - \\ 
		\textbf{2018} & \phantom{0}6\%\! 168 & \phantom{0}4\%\! 185 & \phantom{0}\textcolor{red}{3\%\! 186} & \phantom{0}6\%\! 181 & - & \phantom{0}7\%\! 129 & - & \phantom{0}4\%\! 99\phantom{0} & \phantom{0}4\%\! 120 & - & - & - \\ 
		\textbf{2019} & 10\%\! 125 & \phantom{0}4\%\! 190 & \phantom{0}\textcolor{red}{3\%\! 184} & \phantom{0}7\%\! 177 & \phantom{0}5\%\! 147 & \phantom{0}5\%\! 140 & - & - & - & - & \phantom{0}4\%\! 161 & - \\ 
		\textbf{2021} & 11\%\! 134 & \phantom{0}3\%\! 187 & \phantom{0}\textcolor{red}{2\%\! 195} & \phantom{0}6\%\! 160 & \phantom{0}3\%\! 159 & \phantom{0}4\%\! 139 & \phantom{0}3\%\! 180 & \phantom{0}4\%\! 187 & \phantom{0}6\%\! 120 & \phantom{0}3\%\! 111 & \phantom{0}4\%\! 166 & \phantom{0}4\%\! 111 \\ 
		\textbf{2022} & \phantom{0}\textcolor{red}{3\%\! 216} & \phantom{0}4\%\! 164 & \phantom{0}3\%\! 207 & \phantom{0}7\%\! 150 & \phantom{0}3\%\! 172 & \phantom{0}3\%\! 133 & \phantom{0}2\%\! 186 & \phantom{0}3\%\! 179 & \phantom{0}6\%\! 137 & \phantom{0}7\%\! 99\phantom{0} & - & - \\ 
		\textbf{2023} & \phantom{0}\textcolor{red}{3\%\! 205} & \phantom{0}3\%\! 139 & \phantom{0}6\%\! 86\phantom{0} & \phantom{0}6\%\! 148 & - & \phantom{0}4\%\! 140 & - & - & - & \phantom{0}6\%\! 102 & - & - \\ \bottomrule
		
	\end{tabular*}
\end{table}

Table~\ref{tab:teams_stability} reports the percentage of time robots were fallen and the average distance covered per player, aggregated by team since 2008. The \textit{fallen time} metric reflects the proportion of \texttt{PlayOn} time during which a robot's height was below 0.25~m. Average distances were calculated similarly to average speeds but normalized by the number of players per team, accounting for variations in team sizes over the years. 

Note that \textit{total distance} and \textit{average speed} are not redundant metrics because the proportion of valid samples per match varies according to different factors. Consequently, a team may achieve a high average speed while covering a low total distance if its players fall frequently or matches are interrupted often. 

Since the soccer field expansion in 2012, FC Portugal had prioritized locomotion stability over speed. In 2019, we introduced the league's first running behaviors \cite{abreu2019learning}, including a sprint at 2.62~m/s with a slow turning rate of \ang{10}/s and a more agile run peaking at 1.52~m/s and \ang{160}/s \cite{Abreu2019_Scientific}. However, faster behaviors proved harder to control, as the team's strategy and path planning were not adapted to the drastic changes, leading to increased collisions, fallen time, and overall instability. In 2021, the team introduced the Sprint-Kick (Section~\ref{sec:sprint_kick}), stabilizing at 3.44~m/s with a maximum turning rate of \ang{33}/s. The fallen time did not increase beyond 11\% because the skill was only used during kickoffs.

Curiously, teams that introduced running behaviors experienced a reduction in the average speed. For FC Portugal (2019) and UT Austin Villa (2023), this was due to instability. In contrast, magmaOffenburg (2022) maintained stability, suggesting their reduced speed and distance were due to deliberate strategy changes.

In 2022, we introduced our new codebase with a dribbling skill peaking at 1.43~m/s. The fallen time was reduced from 11\% to 3\%, despite a significant increase in covered distance. Maximum walking speed improved from 0.88~m/s to 0.93~m/s, combined with unmatched omnidirectional capabilities (see Section \ref{sec:directional_analysis}). Then, in 2023, the fast dribbling skill was removed, as explained in Section \ref{sec:intro}.

\subsubsection{Performance Correlation}

To analyze the significance of the previous metrics on team performance, we calculated the Spearman rank correlation coefficients between these metrics and team rankings across all competitions (see Table~\ref{tab:spearman_corr}). Team rankings, provided in Appendix \ref{app:ranks}, indicate relative positions without reflecting the magnitude of performance differences. Unlike Pearson's correlation, which assumes linear relationships, we used Spearman’s rank correlation to assess the strength and direction of the monotonic relationship between variables based on their ranks. Note that in competitions before 2015, some teams share the same rank. For example, in 2008, there are 7 rank groups, with the last five teams (ranks 17-21) all assigned rank 7 for the correlation calculation.

The average correlation coefficients across all competitions were 0.81 for average speed, 0.60 for maximum speed, 0.81 for covered distance, and \(-\)0.49 for fallen time. As expected, the maximum speed is a weaker predictor of performance compared to average speed or covered distance. Notably, in 2019 and 2021, FC Portugal’s maximum speed was significantly higher than other teams, yet it ranked 6th in both competitions.

The fallen time metric shows a negative correlation, indicating that greater fallen times, which suggest lower stability, are associated with lower rankings. While the magnitude of this correlation is weaker compared to other metrics, it still points to a meaningful relationship. The 2012 competition, however, was an outlier, where this trend was inverted, with teams exhibiting higher fallen times ranking better.

\begin{table}[t]
	\caption{Spearman rank correlation coefficients between team rankings and their performance metrics across years}
	\label{tab:spearman_corr}
	\fontsize{7.5pt}{9pt}\selectfont 
	\setlength\tabcolsep{1pt} 
	\begin{tabular*}{\textwidth}{@{\extracolsep\fill}lrrrrrrrrrrrrrrr}
		\toprule
		&   2008 &   2009 &   2010 &   2011 &  2012 &   2013 &   2014 &   2015 &   2016 &   2017 &   2018 &   2019 &   2021 &   2022 &   2023 \\
		\midrule
		Avg. Sp. &  0.81 &  0.72 &  0.86 &  0.86 &  0.70 &  0.90 &  0.79 &  0.79 &  0.87 &  0.75 &  0.79 &  0.82 &  0.92 &  0.79 &  0.77 \\
		Max. Sp. &  0.74 &  0.58 &  0.69 &  0.60 &  0.38 &  0.62 &  0.74 &  0.52 &  0.50 &  0.39 &  0.86 &  0.43 &  0.76 &  0.61 &  0.66 \\
		Distance &  0.80 &  0.89 &  0.88 &  0.89 &  0.88 &  0.89 &  0.65 &  0.80 &  0.88 &  0.85 &  0.86 &  0.89 &  0.85 &  0.78 &  0.31 \\
		Fallen T.  & -0.27 & -0.32 & -0.41 & -0.46 &  0.21 & -0.29 & -0.62 & -0.77 & -0.50 & -0.53 & -0.61 & -0.82 & -0.54 & -0.62 & -0.83 \\
		\bottomrule
	\end{tabular*}
\end{table}

\subsubsection{Directional Analysis}\label{sec:directional_analysis}

In RoboCup competitions, robots mostly walk forward; however, walking backward or sideways is sometimes required, such as when aligning for a shot. To better characterize how teams employ omnidirectional locomotion, we examined how players move in 8 directions: forward, backward, sideways left and right, and the intermediate directions.

Measurements were taken by analyzing 1-second windows where speeds exceeded 0.1 m/s --- below this threshold, intentional direction is hard to determine. Displacement vectors were compared to the robot's average orientation during each window to assign one of 8 movement directions. 

\begin{figure}[t]
	\centering
	\includegraphics[trim={0 0 0 0}, width=1\columnwidth]{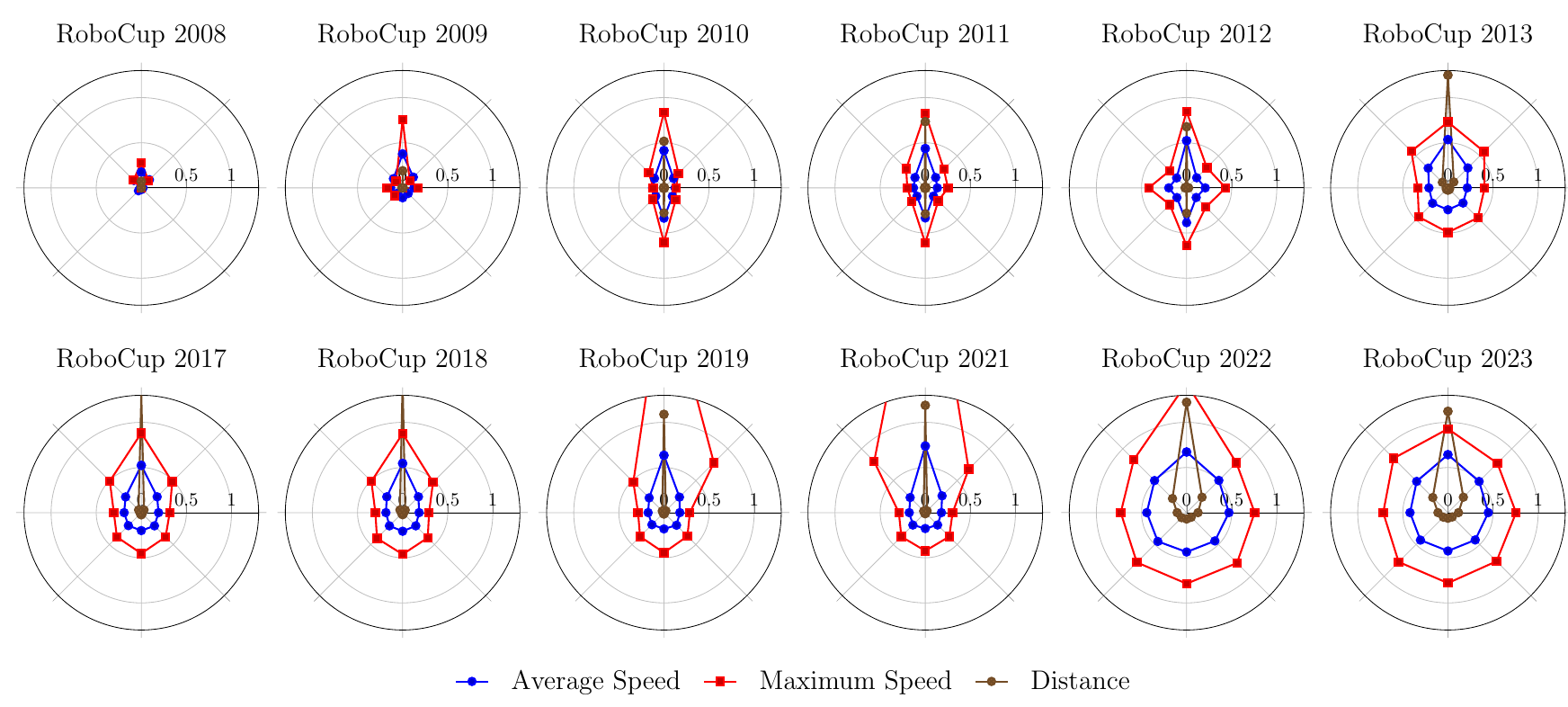}
	
	\caption{FC Portugal locomotion statistics for periods of notable change (2008–2013, 2017–2023). Metrics include average speed (m/s), maximum speed (m/s), and distance covered (hm), plotted radially to show variation by movement direction, relative to the robot. Values at the top represent forward movement, values on the left indicate sideways movement to the left, and so on.}
	\label{fig:radial_FCP}	
\end{figure}

Fig.~\ref{fig:radial_FCP} shows directional metrics for FC Portugal's locomotion during periods of significant evolution (2008–2013, 2017–2023): average speed (m/s), maximum speed (m/s), and distance covered (hm). The radial plots represent a top view of the robot, with the robot's forward direction pointing upward. Data points are positioned radially to represent the direction of movement: forward at the top, backward at the bottom, leftward at the left, and so on.

Initially, the team focused on developing a forward-walking skill, gradually introducing omnidirectional capabilities. Faster forward-running skills were introduced in 2019 and 2021. Asymmetries in the maximum speed arise from biases in these skills, particularly when stopping from a running state \cite{abreu2019learning}. In 2022, the maximum forward speed was influenced by the dribbling skill. However, in 2023 that metric was solely dictated by the omnidirectional walk, revealing its consistent performance in all directions. Moreover, the distance metric indicates that omnidirectional walking was scarcely used until 2022, with back-walking between 2010 and 2012 being an exception. 

\begin{figure}[t]
	\centering
	\includegraphics[trim={0 0 0 0}, width=1\columnwidth]{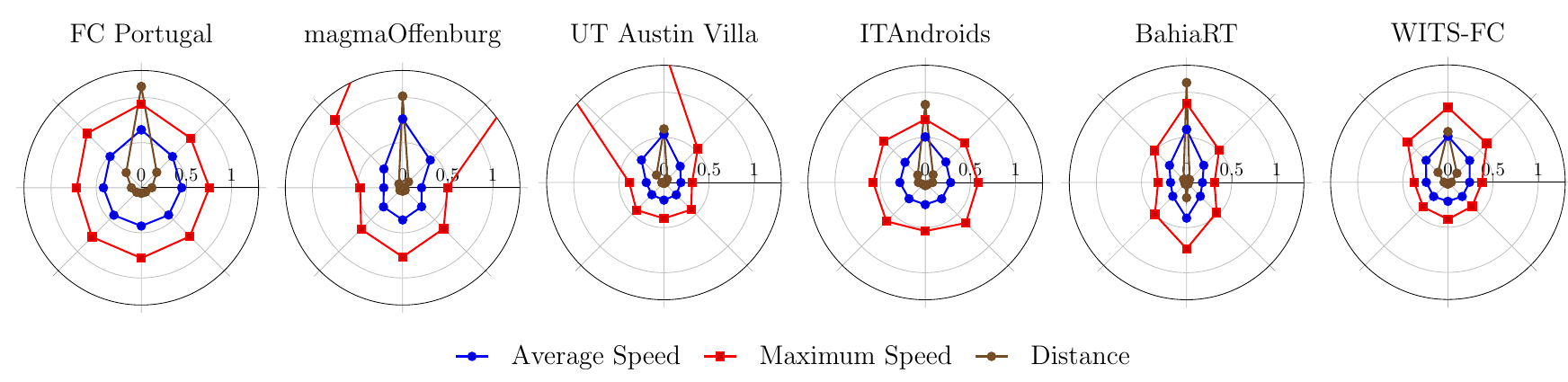}
	
	\caption{3DSSL locomotion statistics for the six teams that participated in RoboCup 2023, ordered by competition ranking. Metrics include average speed (m/s), maximum speed (m/s), and distance covered (hm), plotted radially to show variation by movement direction, relative to the robot. Values at the top represent forward movement, values on the left indicate sideways movement to the left, and so on.}
	\label{fig:radial_2023}	
\end{figure}

\begin{figure}[t]
	\centering
	\includegraphics[trim={0 0 0 0}, width=1\columnwidth]{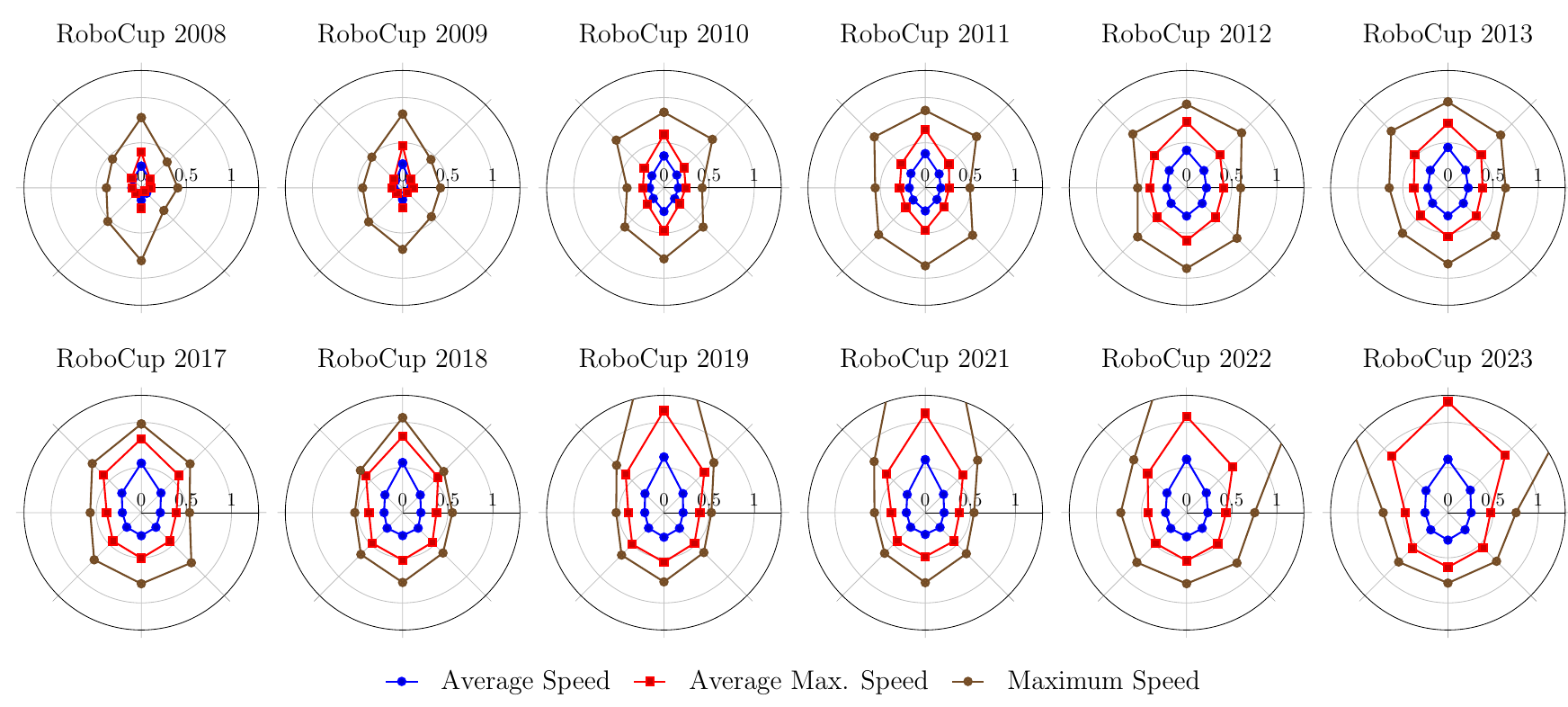}
	
	\caption{3DSSL aggregated locomotion statistics for periods of notable change (2008–2013, 2017–2023). Metrics include average speed, average maximum speed, and maximum speed (all in m/s), plotted radially to show variation by movement direction, relative to the robot. Values at the top represent forward movement, values on the left indicate sideways movement to the left, and so on.}
	\label{fig:radial_all}	
\end{figure}

Fig.~\ref{fig:radial_2023} illustrates the same metrics for the six teams participating in the 2023 competition, arranged in order of their rankings; and Fig.~\ref{fig:radial_all} aggregates speed metrics for all teams over the years, including average speed, average maximum speed, and maximum speed (all in m/s). 

In 2023, the two fastest teams in forward speed were magmaOffenburg and UT Austin Villa. Their skill asymmetries (Fig.~\ref{fig:radial_2023}) are apparent in replays \cite{Glaser_Archive}. Interestingly, these asymmetries are in opposite directions, complementing each other to create a symmetrical pattern in the aggregate plot (Fig.~\ref{fig:radial_all}).

Although the maximum speed metric in the 3DSSL shows minimal apparent improvement from 2008 to 2018, robots fell significantly more often during earlier competitions, limiting the sustainability of their top speed and, consequently, their effectiveness. This conclusion becomes apparent when comparing the average speed metric over the same period.

\section{Codebase Release}

\begin{figure}[!t]
	\centering
	\includegraphics[trim={0 0 0 0}, width=1\columnwidth]{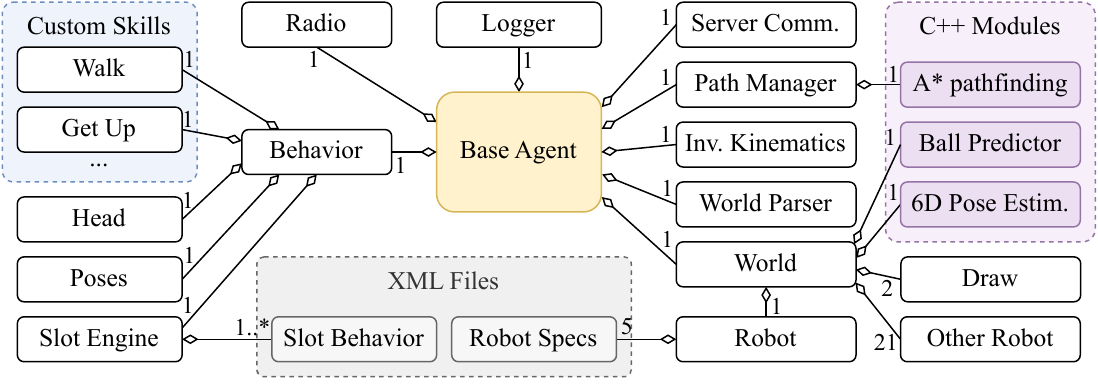}
	\caption{\texttt{Base Agent} structure}
	\label{fig:codebase}	
\end{figure}

Our codebase release is accessible on GitHub\footnote{FC Portugal Codebase release: \url{https://github.com/m-abr/FCPCodebase}}. The project includes a fully functional team of agents that follow a simple formation and attempt to score by kicking the ball toward the opponent's goal. The primary agent derives from \texttt{Base Agent}, a class that provides multiple features, as depicted in Fig.~\ref{fig:codebase}. This class incorporates a representation of both the \texttt{Robot} and the \texttt{World}. The initial representation is obtained from the server and is further enhanced through internal processing, through a \texttt{6D Pose Estimator} \cite{abreu20216d} and a rolling \texttt{Ball Predictor}. The agent is equipped with essential tools such as a pathfinding algorithm, inverse kinematics, and extensive data about the robot and its environment. Team communication is facilitated through the Radio, automatically sharing information about visible players and the ball. While not all skills are included for competitive reasons, this paper provides all the necessary details to train and replicate the  skills mentioned in Table~\ref{tab:results}. The codebase release contains the Omnidirectional Walk, and Dribble v1, along with the underlying Step Baseline skill-set-primitive. Additionally, it includes the latest get-up behaviors, a basic kick, and a basic goalkeeper dive.

\section{Conclusion}

This work introduces the skill set developed for FC Portugal, a team that has proven its capabilities by winning the last two RoboCup 3D Soccer Simulation League World Championships. The methodology used to develop these skills is based on two concepts: a custom algorithm that extends PPO with symmetry-leveraging abilities and skill-set-primitives. These primitives are a subset of motion primitives, and capture common action patterns, making it easier to transition between different behaviors. Furthermore, they allow the policy to be simplified, improving sample efficiency and stability, while still retaining excellent performance in learning complex behaviors. Skill-set-primitives facilitate faster learning without introducing bias, apart from setting the initial exploration area within the solution space.

The first skill created with this framework was the Sprint-Kick, which involves sprinting toward the ball and executing a powerful kick without slowing down. In 2021, this skill was introduced as the fastest sprint to be demonstrated in official league matches (3.69 m/s forward, and 3.44 m/s while curving). It was also the fastest kick.

Following that, we developed the locomotion set, a collection of skills based on one skill-set-primitive, which now forms the core of FC Portugal's team. This set includes an omnidirectional walk that accelerates, decelerates, rotates, and changes direction seamlessly and automatically, while being controlled by a shallow neural network with a single hidden layer of 64 neurons. The Dribble skill allowed for unprecedented ball control and maneuverability, while the Push skill addressed collision-prone scenarios, increasing ball possession and game stability. We employed virtual agents to streamline the learning process in a multi-agent environment, enabling efficient learning within a reasonable timeframe.

Additionally, we have released the codebase of our team to the robotics community. This open-source approach enables other teams to build upon our work, fostering collaboration and innovation in the realm of robotic soccer.

\subsubsection*{Acknowledgments}
The first author was supported by the Foundation for Science and Technology (FCT) under grant SFRH/BD/139926/2018. Additionally, this research was financially supported by FCT/MCTES (PIDDAC), under projects UIDB/00027/2020 (LIACC) and UIDB/00127/2020 (IEETA).

\bibliography{bibliography}
\bibliographystyle{conf}

\newcommand{\tableHeaderStateSpace}[1]{%
	\label{#1}
	\small
	\begin{tabular}{@{}x{0.9cm}x{1.17cm}p{6.41cm}x{1.9cm}x{1.9cm}@{}}
		\toprule
		Param. name & Data size $\times$32b FP & Description & Symmetry Indices & Symmetry Multiplier \\ \midrule
}

\newcommand{\tableHeaderActionSpace}[1]{%
	\label{#1}
	\footnotesize
	\begin{tabular}{@{}p{1.9cm}p{11.65cm}@{}}
		\toprule
		Stage & Description \\ \midrule
}

\newcommand{\tableHeaderConditions}[1]{%
	\label{#1}
	\footnotesize
	\begin{tabular}{@{}p{1.5cm}p{12.05cm}@{}}
		\toprule
		Section & Description \\ \midrule
}

\newcommand{\tableHeaderHyperFirst}[1]{%
	\label{#1}
	\footnotesize
	\begin{tabular*}{\textwidth}{@{\extracolsep\fill}lll@{}}
		\begin{tabular}[t]{@{}lc@{}}
		\toprule
		Parameter & Value \\ \midrule
}

\newcommand{\tableHeaderHyperMore}{%
	\end{tabular} &
	\begin{tabular}[t]{@{}lc@{}}
	\toprule
	Parameter & Value \\ \midrule
}

\newcommand{\urule}{%
	\\[-0.3ex] 
	\arrayrulecolor{gray!30}\midrule
	\arrayrulecolor{black} 
}

\begin{appendices}
\section{}\label{app:ranks}

This appendix provides additional information about team rankings, configuration parameters for the Step Baseline and skill transitions, and all the parameters and training conditions required to replicate the Sprint-Kick and the skills within the Locomotion Set. Finally, it presents additional results for the Sprint-Kick and a compilation of videos referenced throughout the paper.

\begin{table}[h]
	\caption{Team rankings in the RoboCup 3D Soccer Simulation League competition across the years 2008–2012. Teams sharing the same ranking are shown without a separating line.}
	\label{tab:teams_tanks1}
	\small
	\setlength\tabcolsep{1pt} 
	\begin{tabular*}{\textwidth}{@{\extracolsep\fill}cccccc}
		\toprule
		Rank & 2008 & 2009 & 2010 & 2011 & 2012 \\
		\midrule
		1 & SEU-RedSun & SEU-RedSun & Apollo3D & UT Austin Villa & UT Austin Villa \\ \cmidrule{1-1}\cmidrule{2-2}\cmidrule{3-3}\cmidrule{4-4}\cmidrule{5-5}\cmidrule{6-6}
		2 & WrightEagle & Bold Hearts & Nao Team Humboldt & CIT 3D & RoboCanes \\ \cmidrule{1-1}\cmidrule{2-2}\cmidrule{3-3}\cmidrule{4-4}\cmidrule{5-5}\cmidrule{6-6}
		3 & BATS & AmoiensisNQ & HfutEngine3D & Apollo3D & Bold Hearts \\ \cmidrule{1-1}\cmidrule{2-2}\cmidrule{3-3}\cmidrule{4-4}\cmidrule{5-5}\cmidrule{6-6}
		4 & CZU 3D & HfutEngine3D & Bold Hearts & KylinSky3D & magmaOffenburg \\ \cmidrule{1-1}\cmidrule{2-2}\cmidrule{3-3}\cmidrule{4-4}\cmidrule{5-5}\cmidrule{6-6}
		5 & Apollo3D & FUT-K & SEU-RedSun & Bold Hearts & Apollo3D \\ \cmidrule{1-1}
		6 & Borregos3D & opuCI 3D & FC Portugal & FC Portugal & ODENS \\ \cmidrule{1-1}
		7 & Fantasia & JMU 3D & RoboCanes & RoboCanes & CIT 3D \\ \cmidrule{1-1}
		8 & MRL & FC Portugal & DreamWing3D & SEU-RedSun & FC Portugal \\ \cmidrule{1-1}\cmidrule{2-2}\cmidrule{3-3}\cmidrule{4-4}\cmidrule{5-5}\cmidrule{6-6}
		9 & AmoiensisNQ & Borregos3D & opuCI 3D & OxBlue3D & Karachi Koalas \\ \cmidrule{1-1}
		10 & Parsian & magmaOffenburg & magmaOffenburg & beeStanbul & beeStanbul \\ \cmidrule{1-1}
		11 & NAITO-Horizon & BahiaRT & FUT-K & magmaOffenburg & FUT-K \\ \cmidrule{1-1}\cmidrule{6-6}
		12 & OxBlue3D & NomoFC & beeStanbul & DreamWing3D & SOCIO 3D \\ \cmidrule{1-1}\cmidrule{5-5}
		13 & FC Portugal & MRL & BATS & Nexus 3D & L3M-SIM \\ \cmidrule{1-1}
		14 & HfutEngine3D & Scorpius & UT Austin Villa & HfutEngine3D &  \\ \cmidrule{1-1}
		15 & NomoFC & RoboPUB & Nexus 3D & FUT-K &  \\ \cmidrule{1-1}
		16 & opuhana & BATS & Parsian & Karachi Koalas &  \\ \cmidrule{1-1}\cmidrule{2-2}\cmidrule{3-3}\cmidrule{4-4}
		17 & JMU 3D & BMP & BahiaRT & Nao Team Humboldt &  \\ \cmidrule{1-1}
		18 & Scorpius & Parsian & NomoFC & Rail &  \\ \cmidrule{1-1}\cmidrule{5-5}
		19 & PAT & Persia & RoboPUB & L3M-SIM &  \\ \cmidrule{1-1}
		20 & Strive3D & Nexus 3D & Alzahra & Farzanegan &  \\ \cmidrule{1-1}
		21 & UT Austin Villa & NAITO-StrikerS &  & NomoFC &  \\ \cmidrule{1-1}
		22 &  &  &  & BahiaRT &  \\ \bottomrule
	\end{tabular*}
\end{table}

\begin{table}[h]
	\caption{Team rankings in the RoboCup 3D Soccer Simulation League competition across the years 2013–2017. Teams sharing the same ranking are shown without a separating line.}
	\label{tab:teams_tanks2}
	\small
	\setlength\tabcolsep{1pt} 
	\begin{tabular*}{\textwidth}{@{\extracolsep\fill}cccccc}
		\toprule
		Rank & 2013 & 2014 & 2015 & 2016 & 2017 \\
		\midrule
		1 & Apollo3D & UT Austin Villa & UT Austin Villa & UT Austin Villa & UT Austin Villa \\ \cmidrule{1-1}\cmidrule{2-2}\cmidrule{3-3}\cmidrule{4-4}\cmidrule{5-5}\cmidrule{6-6}
		2 & UT Austin Villa & RoboCanes & FUT-K & FUT-K & magmaOffenburg \\ \cmidrule{1-1}\cmidrule{2-2}\cmidrule{3-3}\cmidrule{4-4}\cmidrule{5-5}\cmidrule{6-6}
		3 & FC Portugal & magmaOffenburg & FC Portugal & FC Portugal & FUT-K \\ \cmidrule{1-1}\cmidrule{2-2}\cmidrule{3-3}\cmidrule{4-4}\cmidrule{5-5}\cmidrule{6-6}
		4 & SEU Jolly & FC Portugal & BahiaRT & BahiaRT & RoboCanes \\ \cmidrule{1-1}\cmidrule{2-2}\cmidrule{3-3}\cmidrule{4-4}\cmidrule{5-5}\cmidrule{6-6}
		5 & Karachi Koalas & FUT-K & Apollo3D & magmaOffenburg & AIUT3D \\ \cmidrule{1-1}\cmidrule{4-4}\cmidrule{5-5}\cmidrule{6-6}
		6 & RoboCanes & BahiaRT & magmaOffenburg & ITAndroids & BahiaRT \\ \cmidrule{1-1}\cmidrule{4-4}\cmidrule{5-5}\cmidrule{6-6}
		7 & magmaOffenburg & SEU Jolly & RoboCanes & Miracle3D & FC Portugal \\ \cmidrule{1-1}\cmidrule{4-4}\cmidrule{5-5}\cmidrule{6-6}
		8 & Bold Hearts & HfutEngine3D & Nexus 3D & KgpKubs & KgpKubs \\ \cmidrule{1-1}\cmidrule{2-2}\cmidrule{3-3}\cmidrule{4-4}\cmidrule{5-5}\cmidrule{6-6}
		9 & PHOTON & Karachi Koalas & CIT 3D & HfutEngine3D & ITAndroids \\ \cmidrule{1-1}\cmidrule{4-4}\cmidrule{6-6}
		10 & HfutEngine3D & L3M-SIM & HfutEngine3D &  & Miracle3D \\ \cmidrule{1-1}\cmidrule{4-4}\cmidrule{6-6}
		11 & ITAndroids & ODENS & ITAndroids &  & Nexus 3D \\ \cmidrule{1-1}\cmidrule{4-4}\cmidrule{6-6}
		12 & Paydar 3D & Mithras3D & Miracle3D &  & HfutEngine3D \\ \cmidrule{1-1}\cmidrule{2-2}\cmidrule{6-6}
		13 & FUT-K &  &  &  & RIC-AASTMT \\ \cmidrule{1-1}
		14 & Mithras3D &  &  &  &  \\ \cmidrule{1-1}
		15 & ODENS &  &  &  &  \\ \cmidrule{1-1}
		16 & L3M-SIM &  &  &  &  \\ \cmidrule{1-1}\cmidrule{2-2}
		17 & BahiaRT &  &  &  &  \\ \bottomrule
	\end{tabular*}
\end{table}


\begin{table}[h]
	\caption{Team rankings in the RoboCup 3D Soccer Simulation League competition across the years 2018–2023}
	\label{tab:teams_tanks3}
	\small
	\setlength\tabcolsep{1pt} 
	\begin{tabular*}{\textwidth}{@{\extracolsep\fill}cccccc}
		\toprule
		Rank & 2018 & 2019 & 2021 & 2022 & 2023 \\
		\midrule
		1 & UT Austin Villa & UT Austin Villa & UT Austin Villa & FC Portugal & FC Portugal \\ \cmidrule{1-1}\cmidrule{2-2}\cmidrule{3-3}\cmidrule{4-4}\cmidrule{5-5}\cmidrule{6-6}
		2 & magmaOffenburg & magmaOffenburg & magmaOffenburg & magmaOffenburg & magmaOffenburg \\ \cmidrule{1-1}\cmidrule{2-2}\cmidrule{3-3}\cmidrule{4-4}\cmidrule{5-5}\cmidrule{6-6}
		3 & FC Portugal & WrightOcean & Apollo3D & UT Austin Villa & UT Austin Villa \\ \cmidrule{1-1}\cmidrule{2-2}\cmidrule{3-3}\cmidrule{4-4}\cmidrule{5-5}\cmidrule{6-6}
		4 & BahiaRT & BahiaRT & WrightOcean & Apollo3D & ITAndroids \\ \cmidrule{1-1}\cmidrule{2-2}\cmidrule{3-3}\cmidrule{4-4}\cmidrule{5-5}\cmidrule{6-6}
		5 & KgpKubs & HfutEngine3D & HfutEngine3D & HfutEngine3D & BahiaRT \\ \cmidrule{1-1}\cmidrule{2-2}\cmidrule{3-3}\cmidrule{4-4}\cmidrule{5-5}\cmidrule{6-6}
		6 & Miracle3D & FC Portugal & FC Portugal & Miracle3D & WITS FC \\ \cmidrule{1-1}\cmidrule{2-2}\cmidrule{3-3}\cmidrule{4-4}\cmidrule{5-5}
		7 & ITAndroids & ITAndroids & Miracle3D & ITAndroids &  \\ \cmidrule{1-1}\cmidrule{4-4}\cmidrule{5-5}
		8 &  &  & ITAndroids & KgpKubs &  \\ \cmidrule{1-1}\cmidrule{4-4}\cmidrule{5-5}
		9 &  &  & BahiaRT & WITS FC &  \\ \cmidrule{1-1}\cmidrule{4-4}\cmidrule{5-5}
		10 &  &  & KgpKubs & BahiaRT &  \\ \cmidrule{1-1}\cmidrule{4-4}
		11 &  &  & WITS FC &  &  \\ \cmidrule{1-1}\cmidrule{4-4}
		12 &  &  & MIRG &  &  \\ \bottomrule
	\end{tabular*}
\end{table}

\clearpage

\subsection{Step Baseline and Transitions} \label{app:parameters}

\begin{table}[h]
	\centering
	\caption{Step Baseline Parameters}
	\label{tab:step_baseline_params}
		\begin{tabular}{@{}lcccc@{}}
			\toprule
			Parameter & \multicolumn{1}{c}{Walk} & \multicolumn{1}{c}{Dribble (v1)} & \multicolumn{1}{c}{Push} & Dribble (v2) \\ \midrule
			$p_y/F_y$ & \multicolumn{1}{c}{1.12} & \multicolumn{1}{c}{1.20 \footnotemark[1]} & \multicolumn{1}{c}{1.20 \footnotemark[1]} & 1.12 \\ \cmidrule{2-5}%
			H (m) & \multicolumn{4}{c}{0.02} \\ 
			T (s) & \multicolumn{4}{c}{0.16} \\ 
			$c_z$ (\%) & \multicolumn{4}{c}{70} \\ \bottomrule
		\end{tabular}
		\footnotetext[1]{except for R3 ($p_y/F_y=0.90$)}
\end{table}

\begin{table}[h]
	\centering
	\caption{Skill transitions}
	\label{tab:transitions}
	\begin{tabular}{lccccc@{}}
		\toprule
		\backslashbox{First}{Second} & Get Up & Walk & Kick & Dribble & Push \\ \midrule
		Get Up & - & Trained & No & No & No \\
		Walk & No & - & Trained & Trained & Trained \\ 
		Kick & No & Innate & - & No & No \\
		Dribble & No & Assisted & No & - & No \\
		Push & No & Assisted & No & Innate & - \\ \bottomrule
	\end{tabular}
\end{table}

\subsection{Sprint-Kick} \label{app:Sprint-Kick}

\begin{table}[h]
	\caption{Sprint-Kick state space: \textit{Sprint Forward}, \textit{Sprint \& Rotate}, \textit{Kick}}
	\tableHeaderStateSpace{tab:obs_srk} 
		step & 1 & Step counter\footnotemark[1] \cite{abreu2019learning} & 0 & 1 \urule
		count & 1 & Counter since last side switch & 0 & 1 \urule
		z & 1 & Head height prediction \cite{abreu2019learning} & 0 & 1 \urule
		gyro & 3 & Gyroscope (x-forward, y-left, z-up) & (0,1,2) & (-1,1,-1) \urule
		dgyro & 3 & $\frac{d}{dt}\begin{bmatrix}
			\text{gyro}_x & \text{gyro}_y & \text{gyro}_z
		\end{bmatrix}$ for $\Delta t=0.02$ s & (0,1,2) & (-1,1,-1) \urule
		acc & 3 & Accelerometer (x-forward, y-left, z-up) & (0,1,2) & (1,-1,1) \urule
		dacc & 3 & $\frac{d}{dt}\begin{bmatrix}
			\text{acc}_x & \text{acc}_y & \text{acc}_z
		\end{bmatrix}$ for $\Delta t=0.02$ s & (0,1,2) & (1,-1,1) \urule
		feet & 12 & (l)eft and (r)ight feet sensors with (p)oint of origin and (f)orce vector: (lpx,lpy,lpz,lfx,lfy,lfz,rpx,rpy,rpz,rfx,rfy,rfz) & (6,7,8,9,10,11, 0,1,2,3,4,5) & (1,-1,1,1,-1,1, 1,-1,1,1,-1,1) \urule
		dfeet & 12 & $\frac{d}{dt}\begin{bmatrix}
			\text{feet}_0 & \ldots & \text{feet}_{11}
		\end{bmatrix}$ for $\Delta t=0.02$ s & (6,7,8,9,10,11, 0,1,2,3,4,5) & (1,-1,1,1,-1,1, 1,-1,1,1,-1,1) \urule
		joints& 20 & All joint positions excluding head and toes & Sagittal reflection\footnotemark[2] & Sagittal reflection\footnotemark[2] \urule
		djoints & 20 & $\frac{d}{dt}\begin{bmatrix}
			\text{joints}_0 & \ldots & \text{joints}_{19}
		\end{bmatrix}$ for $\Delta t=0.02$ s & {Sagittal reflection}\footnotemark[2] & {Sagittal reflection}\footnotemark[2] \urule
		target\footnotemark[3] & 1 & Target direction relative to the robot's torso & 0 & -1 \urule
		ball\footnotemark[4] & 3 & Ball position relative to the robot's torso \quad\quad\quad\quad (x-forward, y-left, z-up) & (0,1,2) & (1,-1,1) \\ \bottomrule
	\end{tabular}
\footnotetext[1]{In \textit{Sprint Forward} and \textit{Sprint \& Rotate}, the counter has an upper limit of 93. A limit prevents the neural network from encountering higher values during testing than during training. The specific value was optimized as a hyperparameter, as it influences the final behavior.} 
\footnotetext[2]{The symmetry operation associated with this sagittal reflection depends on the joints' order}
\footnotetext[3]{The target is only included in \textit{Sprint \& Rotate}}
\footnotetext[4]{The ball position is only included in \textit{Kick}}
\end{table}

\begin{table}[h]
	\caption{\small Sprint-Kick action space modification sequence: \\ \textit{Sprint Forward}, \textit{Sprint \& Rotate}, \textit{Kick}}
	\tableHeaderActionSpace{tab:ac_seq_srk}
		1. Action & Raw target positions for 22 joints (includes toes, excludes  head joints) \\ 
		2. Symmetry & The positions are reflected with respect to the robot's sagittal plane when the left leg is positioned as the front leg (see Section \ref{sec:train_env_sprint_kick})\\ 
		3. IIR Filter & An exponential moving average filter is applied with a smoothing factor of 0.5  \\ 
		4. Map & The raw values are mapped to the position ranges of each individual joint \\ 
		5a. Bias & The initial joint positions are added as a permanent bias (only for \textit{Sprint Forward})\\ 
		5b. Primitive & Add cycle capture primitive (only for \textit{Sprint \& Rotate} and \textit{Kick})  \\ 
		6. Conversion & First, predict current joint positions by combining the last action with previous positions (the server has a 1-step delay). Then, determine joint speeds through proportional control. \\ \bottomrule
	\end{tabular}
\end{table}

\begin{table}[h]
	\caption{Sprint-Kick RL conditions: \textit{Sprint Forward}}
	\tableHeaderConditions{tab:rl_cond_srk_fwd}
		Reset & The robot is reset to a fixed initial pose with bent knees, which is then added as a continuous bias (see Table~\ref{tab:ac_seq_srk}, 5a) \\ 
		Termination & Robot has fallen ($z<0.27$) or timeout (4 s)\\ 
		Episode & There is no specific configuration during the episode  \\ 
		Reward & $r=\frac{d\text{x}}{dt}\,$ for $\Delta t=0.06$ s. There is no penalty for deviations in $y$. The reward is updated every vision cycle. \\ \bottomrule
	\end{tabular}
\end{table}

\begin{table}[h]
	\caption{Sprint-Kick RL conditions: \textit{Sprint \& Rotate}}
	\tableHeaderConditions{tab:rl_cond_srk_rot}
		Reset & \textit{Sprint Forward} is executed for 1.4 s (this parameter was found to be the minimal period for which the resulting behavior, \textit{Sprint \& Rotate},  achieves maximum speed) \\ 
		Termination & Robot has fallen ($z<0.34$) or timeout (7 s)\\ 
		Episode & A new target direction is randomly generated every 0.28 s (cycle caption period, see Fig.~\ref{fig:srk_execution}). The target's range is [-10,10] deg, relative to the current orientation of the robot's torso.  \\ 
		Reward & $r=\text{speed\_2d} \times (1-\text{direction\_error}/10)\,$, with the error given in degrees. The reward is updated every vision cycle. \\ \bottomrule
	\end{tabular}
\end{table}

\begin{table}[h]
	\caption{Sprint-Kick RL conditions: \textit{Kick}}
	\tableHeaderConditions{tab:rl_cond_srk_kick}
		Reset & The ball is placed in a random but reachable position. \textit{Sprint Forward} is executed for 1.4~s. Then, \textit{Sprint \& Rotate} is executed until the ball is within 1 meter. \\ 
		Termination & Robot has fallen ($z<0.33$) or timeout (0.4~s)\\ 
		Episode & There is no specific configuration during the episode \\ 
		Reward & $r=\text{direction\_unit\_vector} \cdot \text{ball\_velocity\_2d}\,$, where $\cdot$ denotes the dot vector. The direction vector is determined immediately after reset by evaluating the direction of the ball relative to the robot. The reward is given at the terminal step.\\ \bottomrule
	\end{tabular}
\end{table}

\begin{table}[h]
	\caption{\small Sprint-Kick hyperparam.: \textit{Sprint Forward} (SF), \textit{Sprint \& Rotate} (SR), \textit{Kick} (K)}
	\tableHeaderHyperFirst{tab:hypers_srk}
		batch steps (K) & 256  \\ 
		batch steps (SF, SR) & 1024  \\ 
		clip range & 0.2 \\ 
		entropy coefficient & 0 \\ 
		environments & 16 \\ 
		epochs & 10 \\ 
		GAE $\lambda$ (lambda) & 0.95 \\ \bottomrule
	\tableHeaderHyperMore
		$\gamma$ (gamma) & 0.99 \\ 
		learning rate (LR) & \num{3e-4} \\ 
		LR scheduler (SF) & constant \\ 
		LR scheduler (SR, K) & linear \\ 
		mini-batch size & 64  \\ 
		NN activation & ReLU  \\ \bottomrule	
	\tableHeaderHyperMore
		NN policy arch. & [64,64] \\ 
		NN value arch. & [64,64] \\ 	
		total time steps (K) & 40M \\ 
		total time steps (SF) & 200M \\ 
		total time steps (SR) & 100M \\ 
		value coefficient & 0.5 \\ \bottomrule
\end{tabular}\end{tabular*}\end{table}

\clearpage

\subsection{Locomotion set} \label{app:locomotion_set}

\begin{table}[h]
	\caption{\textit{Omnidirectional Walk} state space}
	\tableHeaderStateSpace{tab:obs_walk}
		step & 1 & Step counter (max: 120)\footnotemark[1] \cite{abreu2019learning} & 0 & 1 \urule 
		z & 1 & Head height\footnotemark[2]  & 0 & 1 \urule 
		dz & 1 & $\frac{d\text{z}}{dt}\,$ for $\Delta t=0.04$ s & 0 & 1 \urule 
		roll & 1 & Torso roll angle\footnotemark[2]  & 0 & -1 \urule 
		pitch & 1 & Torso pitch angle\footnotemark[2]  & 0 & 1 \urule 
		gyro & 3 & Gyroscope (x-forward, y-left, z-up) & (0,1,2) & (-1,1,-1) \urule 
		acc & 3 & Accelerometer (x-forward, y-left, z-up) & (0,1,2) & (1,-1,1) \urule 
		feet & 12 & (l)eft and (r)ight feet sensors with (p)oint of origin and (f)orce vector: (lpx,lpy,lpz,lfx,lfy,lfz,rpx,rpy,rpz,rfx,rfy,rfz) & (6,7,8,9,10,11, 0,1,2,3,4,5) & (1,-1,1,1,-1,1, 1,-1,1,1,-1,1) \urule 
		ankles & 6 & (l)eft and (r)ight ankle position relative to the robot's hip: (lx,ly,lz,rx,ry,rz) & (3,4,5,0,1,2) & (1,-1,1,1,-1,1) \urule 
		feetori & 6 & (l)eft and (r)ight feet orientation relative to the robot's hip: (lx,ly,lz,rx,ry,rz) & (3,4,5,0,1,2) & (-1,1,-1,-1,1,-1)\!\! \urule 
		shldrs & 4 & (l)eft and (r)ight shoulder joints positions & Sagittal reflection\footnotemark[3] & Sagittal reflection\footnotemark[3] \urule 
		action & 16 & Action array generated in last time step & Sagittal reflection\footnotemark[3] & Sagittal reflection\footnotemark[3] \urule 
		sbprog & 1 & Step Baseline cycle progress $\in[0,1]$ & 0 & 1 \urule 
		sbleg & 2 & One-hot encoding for the current Step Baseline swing leg (left leg, right leg)  & (1,0) & (1,1) \urule 
		targp & 2 & Filtered\footnotemark[4] 2D target position vector, relative to the robot's torso (x-forward, y-left) & (0,1) & (1,-1) \urule 
		targv & 2 & $\frac{d}{dt}\begin{bmatrix}
			\text{targp}_x & \text{targp}_{y}
		\end{bmatrix}$ for $\Delta t=0.02$ s & (0,1) & (1,-1) \urule
		targdir & 1 & Filtered\footnotemark[4] target direction, relative to the robot's torso & 0 & -1 \\ \bottomrule
	\end{tabular}
	\footnotetext[1] {A limit prevents the neural network from encountering higher values during testing than during training. The value 120 was optimized as a hyperparameter, as it influences the final behavior.}
	\footnotetext[2] {Values extracted from 6D pose \cite{abreu20216d}}
	\footnotetext[3] {The symmetry operation associated with this sagittal reflection depends on the joints' order}
	\footnotetext[4] {targp is limited to a magnitude of 0.5 m and a maximum variation of 0.014 m per time step (0.02~s), and targdir is limited to 45 deg and a maximum variation of 1.6 deg per time step}
\end{table}

\begin{table}[h]
	\caption{\textit{Omnidirectional Walk} action space modification sequence}
	\tableHeaderActionSpace{tab:ac_seq_walk}
		1. Action & Relative 6D pose of ankles relative to hip (12 values), shoulder joint positions (4 values)  \\ 
		2. IIR Filter & An exponential moving average filter is applied with a smoothing factor of 0.2  \\ 
		3. Global scale & The filtered action is multiplied by $\;\| \text{targp} \|_2 \times 3.5 + 0.3\;$ if $\;\| \text{targp} \|_2<0.2\;$ to encourage the robot to minimize movement as it approaches the target position  \\ 
		4. Map & The individual values are mapped to the desired initial exploration range \\ 
		5. Primitive & Add Step Baseline primitive \\ 
		6. Constraints & Limit leg joints to prevent  self-collisions \\ 
		7. IK & Apply inverse kinematics to obtain the target joint positions \\ 
		8. Conversion & First, predict current joint positions by combining the last action with previous positions (the server has a 1-step delay). Then, determine joint speeds through proportional control. \\ \bottomrule
	\end{tabular}
\end{table}

\begin{table}[h]
	\caption{\textit{Omnidirectional Walk} RL conditions}
	\tableHeaderConditions{tab:rl_cond_walk}
		Reset & Execute Get Up behavior \\ 
		Termination & Robot has fallen ($z<0.35$)\\ 
		Episode & Two targets are created: position (TP) and direction (TD). A random velocity is periodically assigned (avg. interval of 1.6 s) to a moving TP within a work area of 196 $\text{m}^2$. Upon leaving the work area, the TP is relocated to its center. In 30\% of episodes, TD coincides with TP, whereas in 70\% of episodes, TD changes randomly at an avg. interval of 4 s. \\ 
		Reward & $r=\text{progress} \times \text{direction\_penalty} + \text{idle\_bonus} \,$, where progress represents the distance reduction in relation to TP, $\text{direction\_penalty}=1.03^{-\text{direction\_error}}$ is a penalty for deviating direction\_error degrees from TD, and idle\_bonus is an incentive for precision, rewarding short movements when the target is within 0.2 m. The reward is updated every vision cycle. \\ \bottomrule
	\end{tabular}
\end{table}

\begin{table}[h]
	\caption{\textit{Omnidirectional Walk} hyperparameters}
	\tableHeaderHyperFirst{tab:hypers_walk}
		batch steps & 2048  \\ 
		clip range & 0.2 \\ 
		entropy coefficient & 0 \\ 
		environments & 32 \\ 
		epochs & 10 \\ 
		GAE $\lambda$ (lambda) & 0.95 \\ \bottomrule
	\tableHeaderHyperMore
		$\gamma$ (gamma) & 0.99 \\ 
		learning rate (LR) & \num{3e-4} \\ 
		LR scheduler & constant \\ 
		mini-batch size & 64  \\ 
		NN activation & ReLU  \\ 	
		NN policy arch. & [64] \\ \bottomrule
	\tableHeaderHyperMore
		NN value arch. & [64] \\ 	
		PSL value weight & 0.5 \\ 
		PSL policy weight & 0.005 \\ 	
		total time steps & 50M \\ 
		value coefficient & 0.5 \\ \bottomrule
\end{tabular}\end{tabular*}\end{table}

\begin{table}[h]
	\caption{Kick state space: \textit{Short Kick} and \textit{Long Kick}}
	\label{tab:obs_kick}
		\footnotesize
		\begin{tabular}{x{1.54cm}x{1.62cm}p{9.545cm}}
			\toprule
			Param. name & Data size $\times$32b FP & Description \\ \midrule
			step & 1 & Step counter\footnotemark[1] \cite{abreu2019learning} \urule 
			z & 1 & Head height\footnotemark[2]  \urule 
			dz & 1 & $\frac{d\text{z}}{dt}\,$ for $\Delta t=0.04$ s \urule 
			roll & 1 & Torso roll angle\footnotemark[2]  \urule 
			pitch & 1 & Torso pitch angle\footnotemark[2]  \urule 
			gyro & 3 & Gyroscope (x-forward, y-left, z-up) \urule 
			acc & 3 & Accelerometer (x-forward, y-left, z-up) \urule 
			feet & 12 & (l)eft and (r)ight feet sensors with (p)oint of origin and (f)orce vector:  (lpx,lpy,lpz,lfx,lfy,lfz,rpx,rpy,rpz,rfx,rfy,rfz) \urule 
			joints & 16 & All joint positions excluding head, toes and 4 elbow joints \urule 
			djoints & 16 & $\frac{d}{dt}\begin{bmatrix}
				\text{joints}_0 & \ldots & \text{joints}_{15}
			\end{bmatrix}$ for $\Delta t=0.02$ s \urule 
			ballp & 3 & Ball position relative to the robot's hip (x-forward, y-left, z-up) \urule 
			ballv & 3 & $\frac{d}{dt}\begin{bmatrix}
				\text{ball}_x & \text{ball}_y & \text{ball}_z
			\end{bmatrix}$ for $\Delta t=0.04$ s \urule 
			balld & 1 & Euclidean distance from ball to robot's hip \urule 
			targdir & 1 & Target direction, relative to the robot's torso \urule 
			targd \footnotemark[3] & 1 & Target distance \\ \bottomrule
		\end{tabular}
		
	\footnotetext[1] {Step counter limits are not needed when testing episodes are never longer than training episodes}
	\footnotetext[2] {Values extracted from 6D pose \cite{abreu20216d}}
	\footnotetext[3]{The target distance is only included in \textit{Short Kick}}
\end{table}

\begin{table}[h]
	\caption{Kick action space modification sequence: \textit{Short Kick} and \textit{Long Kick}}
	\tableHeaderActionSpace{tab:ac_seq_kick}
		1. Action & Raw target speed for all joints, excluding head, toes and 4 elbow joints (16 values)  \\ 
		2. Map & The individual values are mapped to the desired initial exploration range \\ 
		3. Bias & An initial bias is added to pull the kicking leg back for 0.04 s (for \textit{Short Kick}) or 0.12 s (for \textit{Long Kick}) to generate a back swing \\ \bottomrule
	\end{tabular}
\end{table}

\begin{table}[h]
	\caption{Kick RL conditions: \textit{Short Kick} and \textit{Long Kick}}
	\tableHeaderConditions{tab:rl_cond_kick}
		Reset & The target direction is constant. The robot is initially placed near the ball with both a random position and orientation. The ball is given a random velocity with a magnitude inferior to 1 m/s. Then, the robot aligns itself with the ball and the target as it approaches. The alignment is stricter for the \textit{Long Kick}. Additionally, for the \textit{Short Kick}, the target distance is randomly set between 3 m and 9 m.  \\ 
		Termination & Timeout (0.20 s for \textit{Short Kick} and 0.32 s for \textit{Long Kick})\\ 
		Episode & There is no specific configuration during the episode \\ 
		Reward & $r=\text{initial\_ball\_targ\_dist} - \text{final\_ball\_targ\_dist}$, where the final distance between ball and target is measured after the episode terminates, when the ball speed drops below 0.2 m/s. The reward waits for the ball to stop but is assigned to the terminal step. \\ \bottomrule
	\end{tabular}
\end{table}

\begin{table}[h]
	\caption{Kick hyperparameters: \textit{Short Kick} (SK) and \textit{Long Kick} (LK)}
	\tableHeaderHyperFirst{tab:hypers_kick}
		batch steps (LK) & 128  \\ 
		batch steps (SK) & 120  \\ 
		clip range & 0.2 \\ 
		entropy coefficient & 0 \\ 
		environments & 24 \\ 
		epochs & 10 \\ \bottomrule
	\tableHeaderHyperMore
		GAE $\lambda$ (lambda) & 0.95 \\ 
		$\gamma$ (gamma) & 0.99 \\ 
		learning rate (LR) & \num{3e-4} \\ 
		LR scheduler & constant \\ 
		mini-batch size (LK) & 64  \\ 
		mini-batch size (SK) & 60  \\ \bottomrule
	\tableHeaderHyperMore
		NN activation & ReLU  \\ 	
		NN policy arch. & [64] \\ 
		NN value arch. & [64] \\ 	
		total time steps (LK) & 25M \\ 
		total time steps (SK) & 15M \\ 
		value coefficient & 0.5 \\ \bottomrule
\end{tabular}\end{tabular*}\end{table}

\begin{table}[h]
	\caption{\textit{Dribble} state space}
	\tableHeaderStateSpace{tab:obs_dribble}
		step & 1 & Step counter (max: 96)\footnotemark[1] \cite{abreu2019learning} & 0 & 1 \urule 
		z & 1 & Head height\footnotemark[2]  & 0 & 1 \urule 
		dz & 1 & $\frac{d\text{z}}{dt}\,$ for $\Delta t=0.04$ s & 0 & 1 \urule 
		roll & 1 & Torso roll angle\footnotemark[2]  & 0 & -1 \urule 
		pitch & 1 & Torso pitch angle\footnotemark[2]  & 0 & 1 \urule 
		gyro & 3 & Gyroscope (x-forward, y-left, z-up) & (0,1,2) & (-1,1,-1) \urule 
		acc & 3 & Accelerometer (x-forward, y-left, z-up) & (0,1,2) & (1,-1,1) \urule 
		feet & 12 & (l)eft and (r)ight feet sensors with (p)oint of origin and (f)orce vector: (lpx,lpy,lpz,lfx,lfy,lfz,rpx,rpy,rpz,rfx,rfy,rfz) & (6,7,8,9,10,11, 0,1,2,3,4,5) & (1,-1,1,1,-1,1, 1,-1,1,1,-1,1) \urule 
		joints & 20 & All joint positions excluding head and toes & Sagittal reflection\footnotemark[3] & Sagittal reflection\footnotemark[3] \urule 
		djoints & 20 & $\frac{d}{dt}\begin{bmatrix}
			\text{joints}_0 & \ldots & \text{joints}_{19}
		\end{bmatrix}$ for $\Delta t=0.02$ s & Sagittal reflection\footnotemark[3] & Sagittal reflection\footnotemark[3] \urule 
		sbprog & 1 & Step Baseline cycle progress $\in[0,1]$ & 0 & 1 \urule 
		sbleg & 2 & One-hot encoding for the current Step Baseline swing leg (left leg, right leg)  & (1,0) & (1,1) \urule 
		sbsin & 1 & Step Baseline oscillation: sin(sbprog $\times \pi$) & 0 & 1 \urule 
		ballp & 3 & Ball position relative to the robot's hip \quad\quad\quad (x-forward, y-left, z-up) & (0,1,2) & (1,-1,1) \urule 
		ballv & 3 & $\frac{d}{dt}\begin{bmatrix}
			\text{ball}_x & \text{ball}_y & \text{ball}_z
		\end{bmatrix}$ for $\Delta t=0.04$ s & (0,1,2) & (1,-1,1) \urule 
		balld & 1 & Euclidean distance from ball to robot's hip & 0 & 1 \urule 
		targdir & 1 & Filtered\footnotemark[4] target direction, relative to the robot's torso & 0 & -1 \urule 
		targv & 1 & $\frac{d\text{targdir}}{dt}\,$ for $\Delta t=0.02$ s & 0 & -1 \\ \bottomrule
	\end{tabular}
	\footnotetext[1] {A limit prevents the neural network from encountering higher values during testing than during training. The value 96 was optimized as a hyperparameter, as it influences the final behavior.}
	\footnotetext[2] {Values extracted from 6D pose \cite{abreu20216d}}
	\footnotetext[3] {The symmetry operation associated with this sagittal reflection depends on the joints' order}
	\footnotetext[4] {targdir is limited to 80 deg and a maximum variation of 20 deg per time step (0.02~s)}
\end{table}

\begin{table}[h]
	\caption{\textit{Dribble} action space modification sequence}
	\tableHeaderActionSpace{tab:ac_seq_dribble}
		1. Action & Relative 6D pose of ankles relative to hip (12 values), shoulder joint positions (4 values)  \\ 
		2. IIR Filter & An exponential moving average filter is applied with a smoothing factor of 0.15  \\ 
		3. Global scale & A global scaling parameter allows the robot to gradually slow down and revert to the Step Baseline primitive for an easy transition to walking  \\ 
		4. Map & The individual values are mapped to the desired initial exploration range \\ 
		5. Primitive & Add Step Baseline primitive \\ 
		6. Constraints & Limit leg joints to prevent  self-collisions \\ 
		7. IK & Apply inverse kinematics to obtain the target joint positions \\ 
		8. Conversion & First, predict current joint positions by combining the last action with previous positions (the server has a 1-step delay). Then, determine joint speeds through proportional control. \\ \bottomrule
	\end{tabular}
\end{table}

\begin{table}[h]
	\caption{\textit{Dribble} conditions}
	\tableHeaderConditions{tab:rl_cond_dribble}
		Reset & The robot is initially placed near the ball with both a random position and orientation. The ball is given a random velocity with a magnitude inferior to 1 m/s. Then, the robot aligns itself with the ball and starts the approach until it is close enough ($<$ 0.25 m). \\ 
		Termination & Robot has fallen ($z < 0.40$), timeout (40 s), ball was lost, or ball cannot be seen\\ 
		Episode & A new target direction is randomly generated on average every 0.33 s \\ 
		Reward & $r=\text{ball\_speed} \times \cos(\text{direction\_error})$, where direction\_error is the deviation, in radians, between the ball velocity vector and the filtered target direction vector used as observation. In the second dribble version, the reward is set to zero when the ball is within 0.115 meters to avoid triggering the new ball-holding foul. The reward is updated every vision cycle. \\ \bottomrule
	\end{tabular}
\end{table}

\begin{table}[h]
	\caption{\textit{Dribble} hyperparameters}
	\tableHeaderHyperFirst{tab:hypers_dribble}
		batch steps & 2048  \\ 
		clip range & 0.2 \\ 
		entropy coefficient & 0 \\ 
		environments & 32 \\ 
		epochs & 10 \\ 
		GAE $\lambda$ (lambda) & 0.95 \\ \bottomrule
	\tableHeaderHyperMore
		$\gamma$ (gamma) & 0.99 \\ 
		learning rate (LR) & \num{3e-4} \\ 
		LR scheduler & constant \\ 
		mini-batch size & 64  \\ 
		NN activation & ReLU  \\ 	
		NN policy arch. & [64] \\ \bottomrule
	\tableHeaderHyperMore
		NN value arch. & [64] \\ 	
		PSL value weight & 0.5 \\ 
		PSL policy weight & 0.002 \\ 	
		total time steps & 150M \\ 
		value coefficient & 0.5 \\ \bottomrule
\end{tabular}\end{tabular*}\end{table}

\begin{table}[h]
	\caption{\textit{Push LL} state space}
	\tableHeaderStateSpace{tab:obs_push_ll}
		step & 1 & Step counter (max: 96)\footnotemark[1] \cite{abreu2019learning} & 0 & 1 \urule 
		z & 1 & Head height\footnotemark[2]  & 0 & 1 \urule 
		dz & 1 & $\frac{d\text{z}}{dt}\,$ for $\Delta t=0.04$ s & 0 & 1 \urule 
		roll & 1 & Torso roll angle\footnotemark[2]  & 0 & -1 \urule 
		pitch & 1 & Torso pitch angle\footnotemark[2]  & 0 & 1 \urule 
		gyro & 3 & Gyroscope (x-forward, y-left, z-up) & (0,1,2) & (-1,1,-1) \urule 
		acc & 3 & Accelerometer (x-forward, y-left, z-up) & (0,1,2) & (1,-1,1) \urule 
		feet & 12 & (l)eft and (r)ight feet sensors with (p)oint of origin and (f)orce vector: (lpx,lpy,lpz,lfx,lfy,lfz,rpx,rpy,rpz,rfx,rfy,rfz) & (6,7,8,9,10,11, 0,1,2,3,4,5) & (1,-1,1,1,-1,1, 1,-1,1,1,-1,1) \urule 
		joints & 20 & All joint positions excluding head and toes & Sagittal reflection\footnotemark[3] & Sagittal reflection\footnotemark[3] \urule 
		djoints & 20 & $\frac{d}{dt}\begin{bmatrix}
			\text{joints}_0 & \ldots & \text{joints}_{19}
		\end{bmatrix}$ for $\Delta t=0.02$ s & Sagittal reflection\footnotemark[3] & Sagittal reflection\footnotemark[3] \urule 
		sbprog & 1 & Step Baseline cycle progress $\in[0,1]$ & 0 & 1 \urule 
		sbleg & 2 & One-hot encoding for the current Step Baseline swing leg (left leg, right leg)  & (1,0) & (1,1) \urule 
		sbsin & 1 & Step Baseline oscillation: sin(sbprog $\times \pi$) & 0 & 1 \urule 
		ballp & 3 & Ball position relative to the robot's hip \quad\quad\quad (x-forward, y-left, z-up) & (0,1,2) & (1,-1,1) \urule 
		ballv & 3 & $\frac{d}{dt}\begin{bmatrix}
			\text{ball}_x & \text{ball}_y & \text{ball}_z
		\end{bmatrix}$ for $\Delta t=0.04$ s & (0,1,2) & (1,-1,1) \urule 
		balld & 1 & Euclidean distance from ball to robot's hip & 0 & 1 \urule 
		targp & 2 & 2D target position vector, relative to the robot's torso (x-forward, y-left) & (0,1) & (1,-1) \urule 
		opp & 2 & Weighted\footnotemark[4] average of closest opponent's visible limbs positions, relative to our robot's torso & (0,1) & (1,-1) \\ \bottomrule
	\end{tabular}
	\footnotetext[1] {A limit prevents the neural network from encountering higher values during testing than during training. The value 96 was optimized as a hyperparameter, as it influences the final behavior.}
	\footnotetext[2] {Values extracted from 6D pose \cite{abreu20216d}}
	\footnotetext[3] {The symmetry operation associated with this sagittal reflection depends on the joints' order}
	\footnotetext[4] {The relative weight for each limb is $w_r=10^{-6d}$, where $d$ is the respective Euclidean distance}
\end{table}

\begin{table}[h]
	\caption{\textit{Push LL} action space modification sequence}
	\tableHeaderActionSpace{tab:ac_seq_push_LL}
		1. Action & Relative 6D pose of ankles relative to hip (12 values), shoulder joint positions (4 values)  \\ 
		2. IIR Filter & An exponential moving average filter is applied with a smoothing factor of 0.2  \\ 
		3. Map & The individual values are mapped to the desired initial exploration range \\ 
		4. Primitive & Add Step Baseline primitive \\ 
		5. Constraints & Limit leg joints to prevent  self-collisions \\ 
		6. IK & Apply inverse kinematics to obtain the target joint positions \\ 
		7. Conversion & First, predict current joint positions by combining the last action with previous positions (the server has a 1-step delay). Then, determine joint speeds through proportional control. \\ \bottomrule
	\end{tabular}
\end{table}

\begin{table}[h]
	\caption{\textit{Push LL} conditions}
	\tableHeaderConditions{tab:rl_cond_push_ll}
		Reset & One robot from our team approaches the ball, aligns itself with it, and begins the approach until close enough. If it falls or gets overtaken by a teammate, the active player is switched, and the procedure is restarted. \\ 
		Termination & Robot has fallen ($z < 0.40$), timeout (8 s), ball was lost, ball cannot be seen, teammate is closer to ball \\ 
		Episode & The opposing team executes 1 time step: the closest opponent to the ball is physically simulated and uses the Omnidirectional Walk to push the ball to our goal. The other 10 opponents adhere to a simplified locomotion model, use FC Portugal's formation, and are represented as 2D points with velocity and acceleration. Our team executes an analogous behavior, except for the current closest player, which is the only player controlled by the policy being learned.  \\ 
		Reward & $r=\min(\text{ball\_speed},0.5) \times \max(0,\cos(\text{direction\_error}))^7$, where direction\_error is the deviation, in radians, between the ball velocity vector and the target position vector generated by \textit{Push HL}. The ball speed is limited to 0.5 m/s to prioritize ball possession and progression over high speed. The reward is updated every vision cycle. \\ \bottomrule
	\end{tabular}
\end{table}

\begin{table}[h]
	\caption{\textit{Push LL} hyperparameters}
	\tableHeaderHyperFirst{tab:hypers_push_ll}
		batch steps & 4096  \\ 
		clip range & 0.2 \\ 
		entropy coefficient & 0 \\ 
		environments & 32 \\ 
		epochs & 10 \\ 
		GAE $\lambda$ (lambda) & 0.95 \\ \bottomrule
	\tableHeaderHyperMore
		$\gamma$ (gamma) & 0.99 \\ 
		learning rate (LR) & \num{3e-4} \\ 
		LR scheduler & constant \\ 
		mini-batch size & 64  \\ 
		NN activation & ReLU  \\ 	
		NN policy arch. & [64] \\ \bottomrule
	\tableHeaderHyperMore
		NN value arch. & [64] \\ 	
		PSL value weight & 0.5 \\ 
		PSL policy weight & 0.001 \\ 	
		total time steps & 320M \\ 
		value coefficient & 0.5 \\ \bottomrule
\end{tabular}\end{tabular*}\end{table}

\begin{table}[h]
	\caption{\textit{Push HL} state space}
	\tableHeaderStateSpace{tab:obs_push_hl}
		radar & 160 & Radial spatial segmentation where each segment indicates the presence of opponents and/or teammates (see Section~\ref{sec:push_hl}) & Sagittal reflection\footnotemark[1] & Sagittal reflection\footnotemark[1] \\ 
		mate & 1 & Closest teammate Euclidean distance & 0 & 1 \\ 
		opp & 1 & Closest opponent Euclidean distance & 0 & 1 \\ 
		targdir & 1 & Last HL target direction, relative to user-defined long-term goal direction  & 0 & 1 \\ \bottomrule
	\end{tabular}
	\footnotetext[1] {The symmetry operation associated with this sagittal reflection depends on the radar's sensor order}
\end{table}

\clearpage

\begin{table}[h]
	\caption{\textit{Push HL} action space modification sequence}
	\tableHeaderActionSpace{tab:ac_seq_push_HL}
		1. Action & Target direction (1 value)  \\ 
		2. Map & The direction is mapped to the range [-90,90] \\ 
		3. Filter & A filtered direction is set to a maximum variation of 30 deg per iteration \\ \bottomrule
	\end{tabular}
\end{table}

\begin{table}[h]
	\caption{\textit{Push HL} conditions}
	\tableHeaderConditions{tab:rl_cond_push_hl}
		Reset & Wait for \textit{Push LL} to reset \\ 
		Termination & Push LL resets and, simultaneously, ball exits learning area ($x \in [-12,12], y \in [-9,9]$) \\ 
		Episode & Push LL is executed 16 times prior to re-observing the environment and running the policy, at a frequency of 3.125 Hz. \\ 
		Reward & $r=\text{clip}(\text{progress*0.15},-0.02,0.02)+\text{clip}(\text{advantage*0.4},-0.3,0.3)$, with $\text{progress}=\text{initial\_ball\_targ\_dist} - \text{final\_ball\_targ\_dist}$, and $\text{advantage}=\text{closest\_opponent\_dist}-\text{closest\_teammate\_dist}$. As in \textit{Push LL}, the objective is to prioritize ball possession and progression over high speed. The reward is updated every 0.32 s. \\ \bottomrule
	\end{tabular}
\end{table}

\begin{table}[h]
	\caption{\textit{Push HL} hyperparameters}
	\tableHeaderHyperFirst{tab:hypers_push_hl}
		batch steps & 512  \\ 
		clip range & 0.2 \\ 
		entropy coefficient & 0 \\ 
		environments & 32 \\ 
		epochs & 10 \\ 
		GAE $\lambda$ (lambda) & 0.95 \\ \bottomrule
	\tableHeaderHyperMore
		$\gamma$ (gamma) & 0.99 \\ 
		learning rate (LR) & \num{3e-4} \\ 
		LR scheduler & constant \\ 
		mini-batch size & 64  \\ 
		NN activation & ReLU  \\ 	
		NN policy arch. & [64,64] \\ \bottomrule
	\tableHeaderHyperMore
		NN value arch. & [64,64] \\ 	
		PSL value weight & 0.5 \\ 
		PSL policy weight & 0.01 \\ 	
		total time steps & 20M \\ 
		value coefficient & 0.5 \\ \bottomrule
\end{tabular}\end{tabular*}\end{table}

\clearpage

\subsection{Additional results} \label{app:results}

\begin{figure}[h]
	\centering
	\includegraphics[trim={0 0 0 0}, width=0.8\columnwidth]{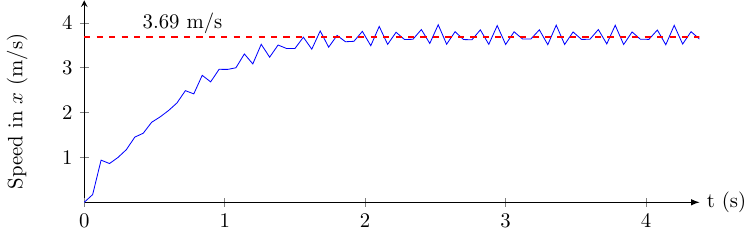}
	\caption{Sprint-Kick: average forward sprint from 5 consecutive samples. The robot stabilizes at an average speed of 3.69 m/s after approximately 2 seconds}
\end{figure}

\begin{figure}[h]
	\centering
	\includegraphics[trim={0 0 0 0}, width=0.9\columnwidth]{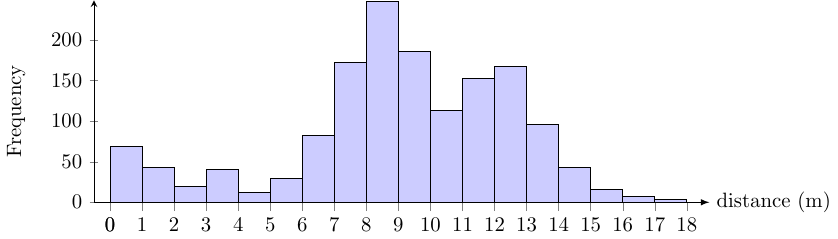}
	\caption{Analysis of 1500 Sprint-Kick samples. The histogram illustrates the distribution of kick distances, i.e., the distance the ball traveled after being kicked. The most frequent range was between 8 and 9 meters.  }
\end{figure}

\subsection{List of videos} \label{app:videos}

\begin{table}[h]
	\vspace{-0.9cm}
	\caption{Demonstration videos}
		\footnotesize
		\begin{tabular}{p{2.68cm}p{2.0cm}p{7.15cm}}
			\toprule
			Skill & Video ID\footnotemark[1] & Description \\ \midrule
			\multirow{3}{*}{Sprint-Kick} & \href{https://youtu.be/Yy1yCM5hwZI}{Yy1yCM5hwZI} & Sprint-Kick training environment \\ 
			 & \href{https://youtu.be/3MND8RVUPBQ}{3MND8RVUPBQ} & Successful Sprint-Kick Kickoffs in RoboCup 2021 \\ 
			 & \href{https://youtu.be/udN2F3oXAec}{udN2F3oXAec} & Sprint-Kick slow motion \\  \midrule
			Omnidirectional Walk & \href{https://youtu.be/dXzIuZlOFZc}{dXzIuZlOFZc}  & Omnidirectional Walk training env. and demonstration \\ \midrule
			Short Kick & \href{https://youtu.be/trn1KzXFhwc}{trn1KzXFhwc}  & Short Kick training environment \\ \midrule
			\multirow{2}{*}{Long Kick} & \href{https://youtu.be/pu4g5wryfEs}{pu4g5wryfEs}  & Best kick goals in RoboCup 2022 \\
			 & \href{https://youtu.be/QQXGgpcnVYE}{QQXGgpcnVYE}  & Long Kick test to assess the final ball position \\ \midrule
			Dribble & \href{https://youtu.be/8UED_Zl-nbQ}{8UED\_Zl-nbQ}  & Dribble evolution, training environment and demonstration \\ \midrule
			\multirow{2}{*}{Push} & \href{https://youtu.be/rGWN83FBdJ4}{rGWN83FBdJ4}  & Push training environment \\ 
			 & \href{https://youtu.be/mhfQ0NAmlMI}{mhfQ0NAmlMI}  & Push highlights from Brazil Open 2022 \\ \bottomrule
		\end{tabular}
	\footnotetext[1] {Access through https://youtu.be/\{Video ID\}}
\end{table}

\end{appendices}

\end{document}